\documentclass[11pt, a4paper, logo, copyright, nonumbering]{xiaomi}

\usepackage[authoryear, sort&compress, round]{natbib}
\usepackage{caption}
\usepackage{subcaption}
\usepackage{xspace}
\usepackage{pifont}
\usepackage{multirow}
\usepackage{makecell}
\usepackage{hyperref}
\usepackage{wrapfig}
\usepackage{graphicx}
\usepackage{amsfonts}
\usepackage{amsmath}
\usepackage{amssymb}
\usepackage{mathtools}
\usepackage{float}
\usepackage{algorithm}
\usepackage{algorithmic}
\usepackage{setspace}
\usepackage{multicol}
\definecolor{xiaomiorange}{HTML}{FF6901}

\title{\centering Stabilizing MoE Reinforcement Learning by Aligning Training and Inference Routers}

\makeatletter
\renewcommand{\author}[1]{\gdef\@author{\Authfont #1}}
\makeatother
\author{%
  \bfseries Wenhan Ma\textsuperscript{1,2,$\ast$} \quad
  Hailin Zhang\textsuperscript{2} \quad Liang Zhao\textsuperscript{2} \quad Yifan Song\textsuperscript{1,2} \\
  \bfseries Yudong Wang\textsuperscript{1,2} \quad
  Zhifang Sui\textsuperscript{1,$\dagger$} \quad
  Fuli Luo\textsuperscript{2,$\dagger$} \\[5pt]
  \normalfont
  \textsuperscript{1}State Key Laboratory of Multimedia Information Processing, \\
  School of Computer Science, Peking University \\
  \textsuperscript{2}LLM Core, Xiaomi
}

\begin{abstract}

Reinforcement learning (RL) has emerged as a crucial approach for enhancing the capabilities of large language models.
However, in Mixture-of-Experts (MoE) models, RL training tends to be unstable and can even collapse entirely.
We analyze the training-inference consistency of MoE models and identify a notable discrepancy in routing behaviors between the two phases.
To address this issue, we propose \textbf{Rollout Routing Replay (R3)}, a novel and effective method that records routing distributions from the inference engine and replays them during training.
R3 significantly reduces training-inference policy KL divergence and mitigates extreme discrepancies without compromising training speed.
Extensive experiments on various settings confirm that R3 succeeds in stabilizing RL training, preventing collapse and outperforming strong baselines. R3 is orthogonal to most policy optimization algorithm improvements, allowing it to be used in conjunction with them.
We believe this work can offer a new solution for stabilizing RL in MoE models.

\end{abstract}

\begin{document}
\maketitle
\renewcommand{\thefootnote}{\fnsymbol{footnote}}
\footnotetext[1]{Work done during an internship at Xiaomi.}
\footnotetext[2]{Co-corresponding authors.}
\renewcommand{\thefootnote}{\arabic{footnote}}

%

\section{Introduction}

Reinforcement learning (RL), especially reinforcement learning with verifiable rewards (RLVR), has become a cornerstone in the post-training of large language models (LLMs)~\citep{ouyang2022training,o1,dpskr1}.
By leveraging large-scale RL, LLMs acquire the advanced capabilities necessary to tackle complex problems, including competition-level mathematics~\citep{dpskr1} and practical code agent tasks~\citep{deepswe2025}, through more profound and extended reasoning.
\begin{figure}[t]
  \centering
  \begin{minipage}[t]{0.72\textwidth}\vspace{0pt}
    \centering
    \includegraphics[width=\textwidth]{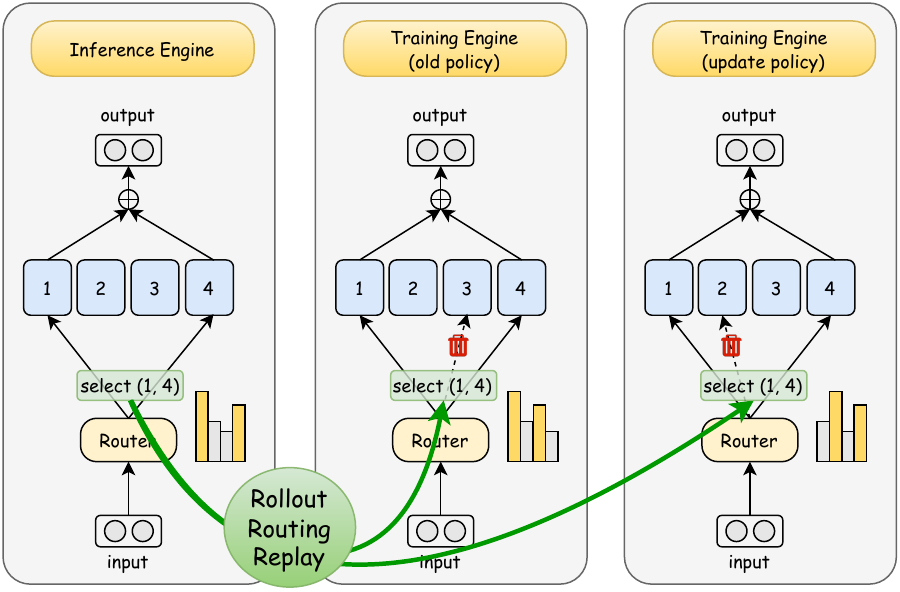}
  \end{minipage}\hfill
  \begin{minipage}[t]{0.26\textwidth} \vspace{0pt}
    \centering
    \begin{minipage}[t]{\linewidth} \vspace{0pt}
      \centering
      \includegraphics[width=\linewidth]{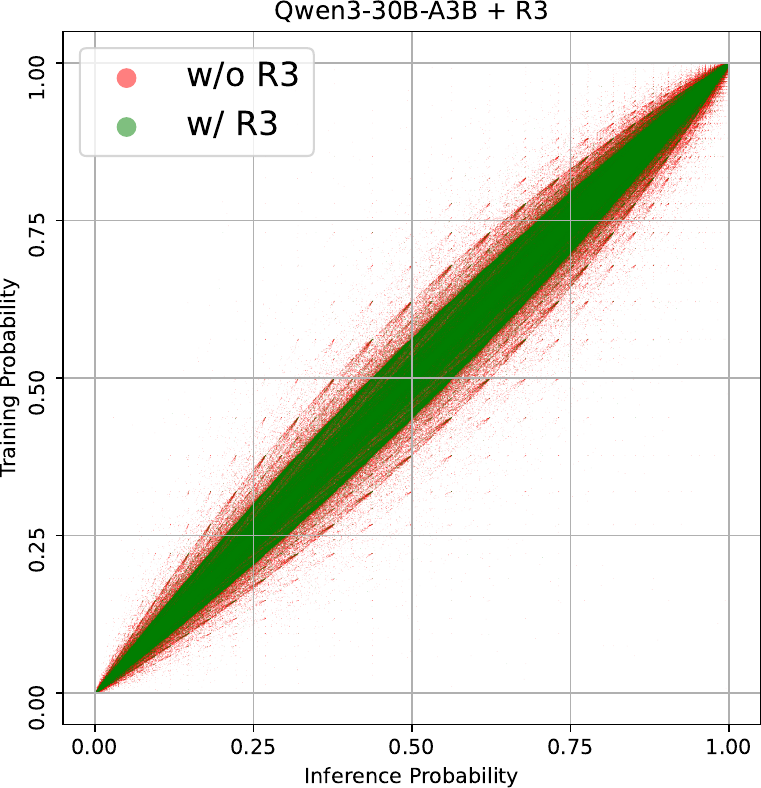}
    \end{minipage}\\[1ex]
    \begin{minipage}[t]{\linewidth}\vspace{0pt}
      \centering
      \includegraphics[width=\linewidth]{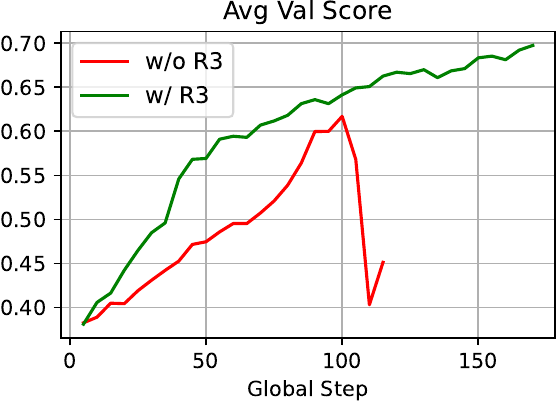}
    \end{minipage}
  \end{minipage}
  \caption{Left: Illustration of the Rollout Routing Replay (R3). Top right: Training and inference discrepancies before and after applying R3. Bottom right: Reinforcement learning training performance before and after applying R3.}
  \label{fig:teaser}
\end{figure}

A key challenge is enhancing the stability of the LLM RL training process.
Modern RL frameworks typically employ distinct engines for inference and training phases (e.g., SGLang~\citep{sglang} for inference and Megatron~\citep{megatron} for training).
This architectural separation can lead to divergent token probabilities, potentially causing catastrophic collapse~\citep{nondeter}.
To mitigate this discrepancy, \citet{tis} incorporate an importance sampling mechanism in policy updating, while \citet{nondeter} introduce specialized compute kernels to reduce nondeterminism during LLM inference.
However, in practice, existing approaches do not fully resolve the intensified off-policy issue that arises during RL training on Mixture-of-Experts (MoE) models.

In this work, we identify the router as a key driver of instability in MoE-based RL. In MoE models, a router dynamically selects and activates a subset of experts for each token. Because routing decisions can vary across inference/training framework, MoE models often exhibit substantially larger policy discrepancies between rollout and optimization than dense models. Rather than relying on heuristics such as filtering out tokens with large discrepancies~\citep{IcePop2025}, we address the problem at its source by explicitly controlling the routing behavior.

To this end, we propose \textbf{Rollout Routing Replay (R3)}, a novel and effective method for stabilizing RL training of MoE models. R3 captures routing distributions from the inference engine during sequence generation and replays them into the training engine. This process reduces the discrepancy between training and inference, leading to a significant reduction in the KL divergence between training and inference policies. By replaying the routing distributions, we can precisely optimize only the experts involved in generating the token, avoiding unnecessary optimization of unused experts that would introduce gradient noise.

We conduct experiments across multiple MoE model families (Qwen3~\citep{qwen3}, DeepSeek-V2-Lite~\citep{deepseekai2024deepseekv2strongeconomicalefficient}), model variants (Base, Instruct-SFT, Reasoning-SFT) and RL tasks (Math, Software engineering). Results show that R3 improves training stability, prevents collapse, and boosts evaluation performance, surpassing strong baseline methods like GSPO and TIS. Speed and memory analysis reveals only a 3.45\% slowdown and manageable memory overhead, demonstrating strong practicality for large-scale training.

We summarize the contributions of our work as follows:
(a) We systematically identify and analyze routing distribution discrepancies between training and inference in MoE models, highlighting their role in training instability.
(b) We propose Rollout Routing Replay, which reuses inference routing distributions inside the training engine to align routing behavior between training and inference.
(c) We apply R3 in multiple settings for MoE reinforcement learning and show that R3 outperforms baseline methods in terms of stability and overall performance.

\section{Preliminaries}

\subsection{Nondeterminism in Modern LLM framework}

Modern LLM training and inference frameworks often exhibit nondeterministic behavior; this is observed not only across different frameworks but also within the same framework. Importantly, such variability is not primarily attributable to mismatched mathematical formulations (e.g., dropout or batch normalization, which are typically not present in modern LLMs). Instead, it arises from numerical effects inherent to finite-precision computation and low-level implementation details. The main contributing factors are as follows:

\textbf{Floating-point addition is non-associative:} Floating-point addition is non-associative, meaning that the result depends on the order of summation. Limited precision and rounding errors can cause discrepancies when the execution order changes, a problem that becomes especially pronounced in iterative computations and parallel reductions.

\textbf{Variation in operator and kernel selection: } High-level frameworks (e.g., PyTorch) may dynamically select different low-level kernels (e.g., CUDA) at runtime based on hardware, batch size, or data type. Although these kernels are mathematically equivalent, they often execute floating-point operations in different orders, which can lead to slight variations in the results.

\textbf{Violation of batch invariance:} As stated in \citep{nondeter}, in asynchronous inference systems, samples may enter and leave dynamically, causing the batch size to vary over time. This can trigger different backend kernel selections, leading to nondeterministic inference results.

\textbf{MoE Router Amplifies Nondeterminism: }
MoE models use a router with a Top-K operation, where tiny numerical differences can change the selected Top-K experts and thus alter the model's computation path. This discrete switch can amplify floating-point perturbations into significant output changes, undermining the stability of reinforcement learning. The proposed R3 method is designed to mitigate this issue.

\subsection{Policy Optimization Algorithm in Reinforcement Learning of LLM}

Recently, reinforcement learning has been employed to train large language models in order to enhance their reasoning capabilities. Popular algorithms in this area include GRPO~\citep{grpo}, DAPO~\citep{yu2025dapo} and GSPO~\citep{gspo}, all of which are derived from the classical Proximal Policy Optimization (PPO)~\citep{ppo} algorithm. However, a critical issue arises from the common practice of using different frameworks for sampling and training: data is sampled via the inference framework, but the loss is computed using the training framework. We denote the output probability distributions of the inference and training frameworks as $\mu$ and $\pi$. Consequently, the actual optimization objective becomes:
\begin{equation}
\label{eq:ppo_objective}
\mathcal{J}_{\text{PPO}}(\theta) = \mathbb{E}_{x \sim \mathcal{D}, y \sim \mu_{\theta_{\text{old}}}(\cdot|x)}
\left[ \frac{1}{|y|} \sum_{t=1}^{|y|} \min \left( w_t\hat{A}_t, \text{clip}(w_t, 1 - \varepsilon, 1 + \varepsilon)\hat{A}_t \right) \right].
\end{equation}
The importance sampling ratio \(w_t\) for the token \(y_t\) in the sequence \(y\) is defined as:

\begin{equation}
\label{eq:importance_ratio}
w_t = \frac{\pi_\theta(y_t|x, y_{<t})}{\pi_{\theta_\text{old}}(y_t|x, y_{<t})}
\end{equation}
This policy discrepancy may drive training instability in reinforcement learning. In MoE models, this problem becomes further exacerbated, as it can result in the loss being computed with experts that were not utilized during sampling, thereby causing more severe policy divergence. Moreover, during parameter optimization, gradients may erroneously propagate to experts that did not participate in sampling, leading to unstable optimization.

\section{Training-Inference Discrepancies in MoE Models}
\label{sec:3}
Training-inference discrepancies in RL frameworks frequently lead to unstable training.
In this section, we demonstrate that this policy discrepancy is significantly amplified in MoE models, primarily stemming from inconsistent routing distributions.
\subsection{Policy Discrepancies Analysis}
\label{sec:3.1}
Here we employ an MoE model, Qwen3-30B-A3B~\citep{qwen3}, for our experiments to analyze the policy discrepancy between the training and inference engines.
First, we use the SGLang inference engine to generate responses for 2,048 mathematics problems, saving the token probabilities for each generated token.
This process yields about 20 million response tokens.
Then these responses are passed through Megatron to obtain the corresponding probabilities assigned by the training engine.
We use several metrics to quantify the divergence between these two probability distributions.

\begin{figure}[t]
\centering
\begin{subfigure}[t]{0.24\textwidth}
    \centering
    \includegraphics[width=\linewidth]{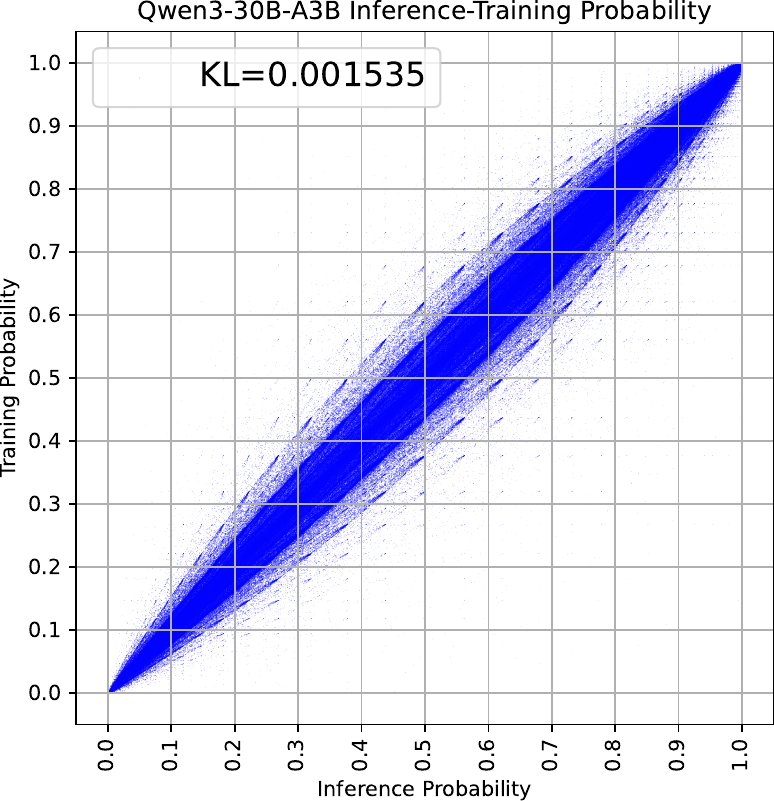}
    \caption{MoE}
    \label{fig:311:a}
\end{subfigure}\hfill
\begin{subfigure}[t]{0.24\textwidth}
    \centering
    \includegraphics[width=\linewidth]{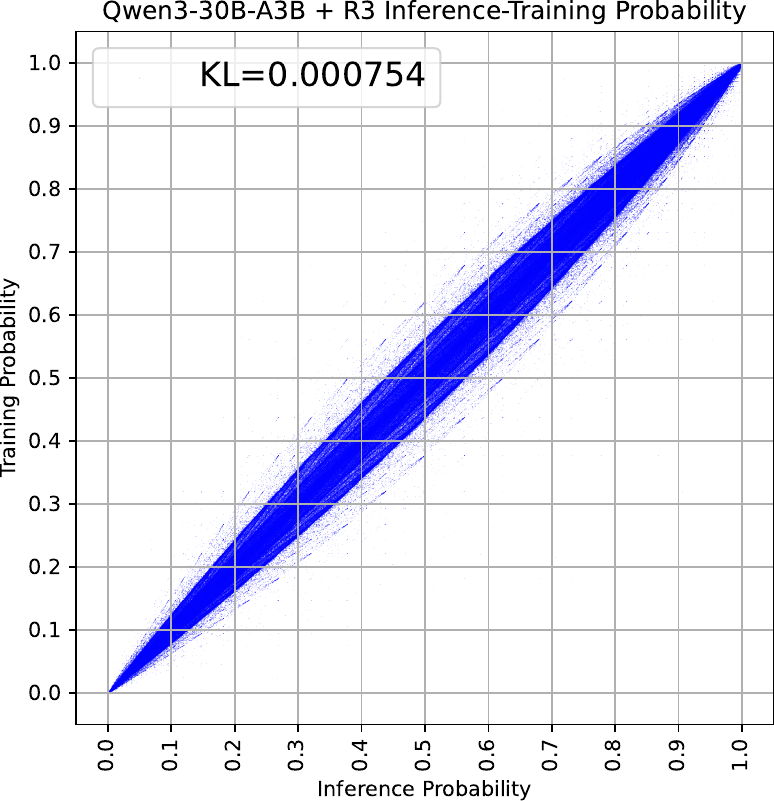}
    \caption{MoE + R3}
    \label{fig:311:b}
\end{subfigure}\hfill
\begin{subfigure}[t]{0.24\textwidth}
    \centering
    \includegraphics[width=\linewidth]{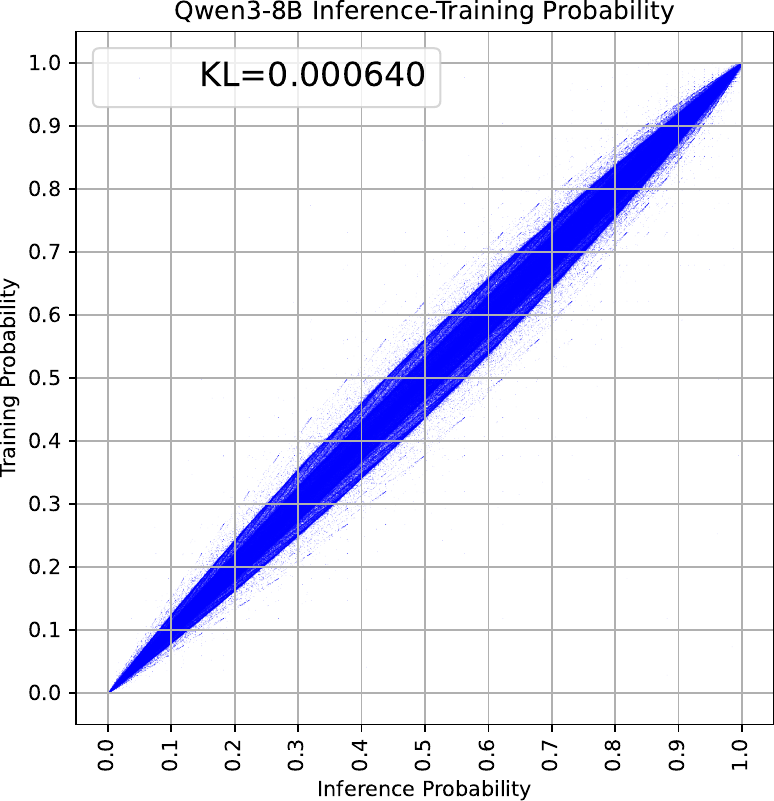}
    \caption{Dense}
    \label{fig:311:c}
\end{subfigure}\hfill
\begin{subfigure}[t]{0.24\textwidth}
    \centering
    \includegraphics[width=\linewidth]{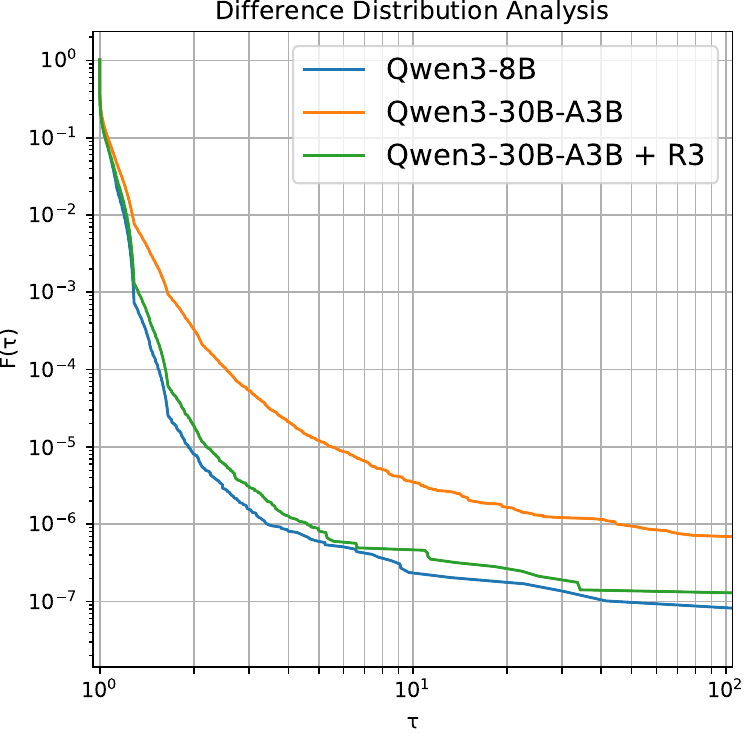}
    \caption{Extreme Token Distribution Analysis}
    \label{fig:311:d}
\end{subfigure}

\caption{(a): Illustration of the training-inference discrepancy in the MoE model.
(b): Illustration of the training-inference discrepancy in the MoE+R3 model.
(c): Illustration of the training-inference discrepancy in the Dense model.
(d): Extreme Token Distribution Function, calculated based on Equation \ref{eq:etdf}.}
\label{fig:311}
\end{figure}

\paragraph{KL Divergence Estimation}
Let $T$ be the set of all response tokens.
We use the \textit{k3} method~\citep{kl} to estimate the KL divergence between the training and inference distributions at token-level. The estimation formula is given as follows:
\begin{equation}
\mathcal{D}_{\text{KL}}(\mu_\theta; \pi_\theta)
\approx \frac{1}{|T|}\sum_{t \in T}\bigl[\frac{\pi_\theta(t)}{\mu_\theta(t)}-1 - \log \frac{\pi_\theta(t)}{\mu_\theta(t)}\bigr].
\label{eq:kl}
\end{equation}
Our calculations show that the estimated KL divergence of Qwen3-30B-A3B (MoE) is $1.535\times10^{-3}$, while that of Qwen3-8B (Dense baseline) is $6.4\times10^{-4}$.

\paragraph{Visualization}
To visualize the training-inference discrepancies of the MoE model, we randomly sample 10 million response tokens and plot a scatter diagram with SGLang probabilities on the x-axis and Megatron probabilities on the y-axis. The degree of concentration around the \(y=x\) line indicates the degree of consistency. As shown in Figure \ref{fig:311:a} and \ref{fig:311:c}, compared to Qwen3-8B, Qwen3-30B-A3B exhibits a much wider scatter band, revealing a larger training-inference discrepancy.

\paragraph{Extreme Token Distribution Analysis}
To quantify the discrepancy between the model's behavior during training and inference, we introduce the Extreme Token Distribution Function, \(F(\tau)\), defined as:
\begin{equation}
    F(\tau) = \frac{1}{|T|} \sum_{t \in T} \mathbf{1}\!\Bigl[\max\Bigl(\frac{\pi_\theta(t)}{\mu_\theta(t)}, \frac{\mu_\theta(t)}{\pi_\theta(t)}\Bigr)>\tau \Bigr].
    \label{eq:etdf}
\end{equation}
This function measures the proportion of \textit{extreme tokens}, i.e., those for which the probability ratio between the training distribution \(\pi_\theta\) and the inference distribution \(\mu_\theta\) surpasses a threshold \(\tau\).
Figure~\ref{fig:311:d} plots this function \(F(\tau)\) against the threshold \(\tau\). The plot reveals that for \(\tau > 2\), the Qwen3-30B-A3B model exhibits a fraction of extreme tokens an order of magnitude larger than that of the Qwen3-8B model. This significant gap indicates a substantially higher degree of training-inference discrepancy in the MoE model.

\subsection{Routing Discrepancies Analysis}
\label{sec:3.2}
\begin{figure}[t]
\centering
\includegraphics[width=\textwidth]{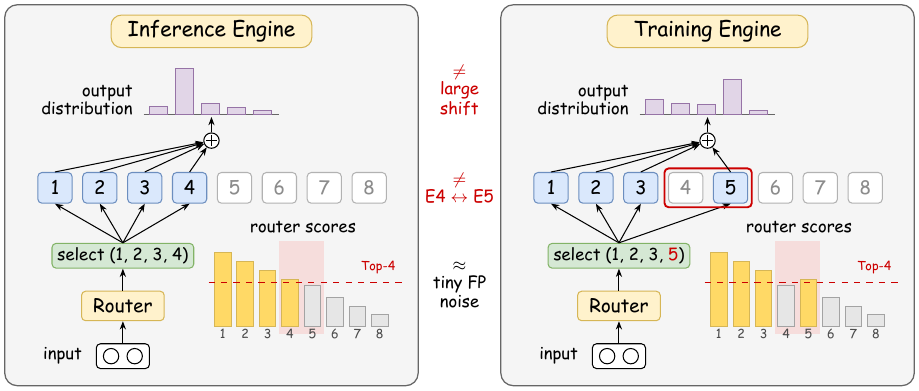}
\caption{Illustration of how the router amplifies a numerical perturbation into a large policy discrepancy.
For the same input token, the router scores in the inference engine (left) and the training engine (right) differ only by floating-point noise. Since experts 4 and 5 are nearly tied at the Top-$K$ cutoff, this tiny difference is enough to flip which of the two is activated.
Because a different expert contributes to the layer output, the resulting output token distribution can differ substantially between the two engines, which makes the collected rollouts off-policy.}
\label{fig:amplify}
\end{figure}

From the perspective of functional continuity, the key difference between MoE and Dense models lies in the non-continuity introduced by routing. In MoE models, small perturbations in the router input can lead to entirely different experts being selected, causing large changes in layer outputs. Dense models, lacking an explicit expert selection process, do not exhibit this phenomenon.
Figure~\ref{fig:amplify} illustrates this amplification effect: when two experts are nearly tied at the Top-$K$ boundary, a floating-point level perturbation is sufficient to flip which expert is activated, and the resulting output token distribution can change substantially.

Based on this, we further analyze router distribution discrepancies between training and inference in MoE models. We use SGLang~\citep{sglang} to generate responses for 2048 mathematical problems with Qwen3-30B-A3B~\citep{qwen3}. For each response, we collect the routing distribution of all tokens (including input tokens) from the inference engine. Then, we feed these sequences into the Megatron engine for forward propagation. This process yields the routing distribution observed from training engine. We compare these two sets of routing information at different levels:
\begin{figure}[!t]
    \centering
\begin{subfigure}[t]{0.30\textwidth}
    \centering
    \includegraphics[width=\textwidth]{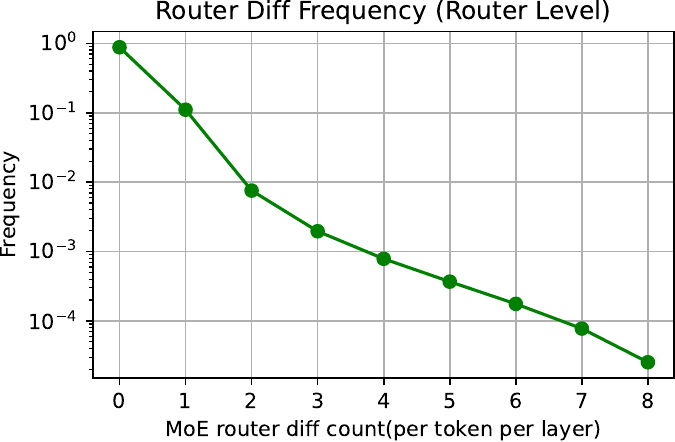}
    \caption{Router-level Difference}
    \label{fig:321:a}
\end{subfigure}
\hfill
\begin{subfigure}[t]{0.30\textwidth}
    \centering
    \includegraphics[width=\textwidth]{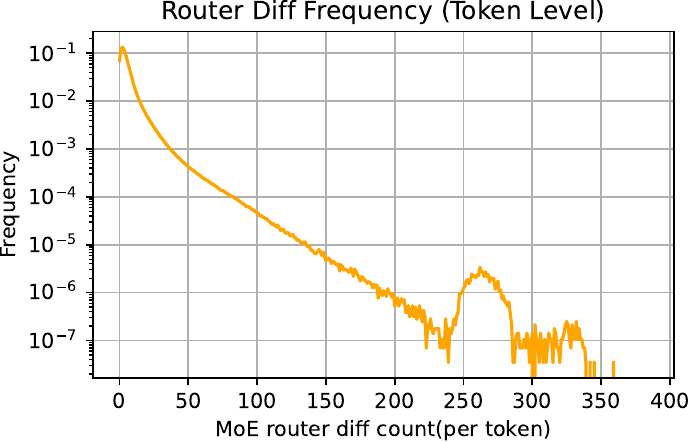}
    \caption{Token-level Difference}
    \label{fig:321:b}
\end{subfigure}
\hfill
\begin{subfigure}[t]{0.30\textwidth}
    \centering
    \includegraphics[width=\textwidth]{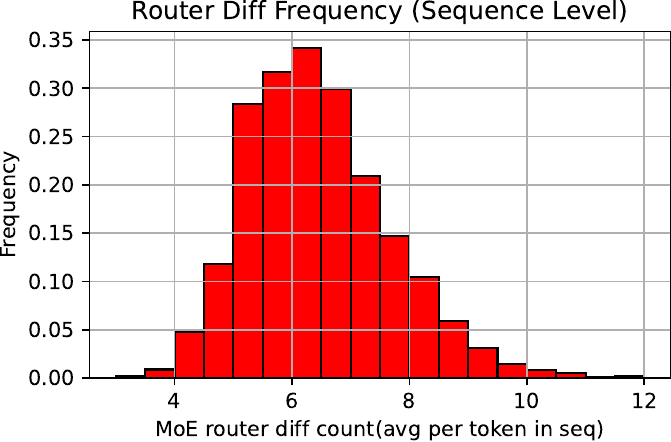}
    \caption{Sequence-level Difference}
    \label{fig:321:c}
\end{subfigure}
\caption{Router discrepancy analysis}
\label{fig:321}
\end{figure}

\textbf{Router-level Comparison}:
  For each token and each MoE layer, we count the number of differing expert choices made by the MoE Router and calculate the frequency of these differences. Figure \ref{fig:321:a} illustrates the results. It is observed that approximately 10\% of the routers select different experts during training compared to inference.

\textbf{Token-level Comparison}:
  For each token, we count the number of differing expert choices made by the MoE Router across all layers. We also determine the frequency of these occurrences. Figure \ref{fig:321:b} presents these findings. It shows that 94\% of tokens select a different expert in at least one layer during the forward pass.

\textbf{Sequence-level Comparison}:
  For a sequence, we compute the router distribution difference of each token, then average over tokens to obtain the mean difference per sequence and plot a histogram (Figure~\ref{fig:321:c}). Results show that the mean difference per token is approximately 6 routers.

These findings demonstrate that during training and inference, MoE models exhibit router distribution discrepancies. In Section~\ref{sec:r3}, we show empirically that routing discrepancies is the main contributor to the additional training-inference discrepancies of MoE compared to Dense models.

\section{Rollout Routing Replay}

\label{sec:r3}

This section provides a detailed description of the implementation of Rollout Routing Replay (R3), its caching support in multi-turn dialogue, and an analysis of its effects on training-inference discrepancies.
\subsection{Implementation}

We take the routing form used in Qwen3~\citep{qwen3} as an example to explain the implementation of R3.
Since R3 only replaces the Top-$K$ expert mask with the one recorded during inference, it applies in the same way to other routing forms.
We first describe the conventional forward pass of a MoE layer within a training framework. Consider a sequence $s$ at the $t$-th token and the MoE layer in the $l$-th Transformer block. Let the input to this layer during training be $\mathbf{x}_{\mathrm{train}}$. The router logits are calculated by:
\begin{equation}
\mathbf{s}_{\mathrm{train}} = \mathbf{x}_{\mathrm{train}} \mathbf{W}_r,
\end{equation}
where $\mathbf{W}_r$ denotes the router's linear weight matrix. Let the number of experts be $M$ and the number of experts to select be $K$.
During training, the router selects the top-$K$ experts based on the logits. This is represented by a binary mask:
\begin{equation}
\mathbf{I}_{\mathrm{train}} = \operatorname{TopKMask}(\mathbf{s}_{\mathrm{train}}, K),
\end{equation}
where $\mathbf{I}_{\mathrm{train}} \in \{0,1\}^M$ and $\sum_i I_{\mathrm{train},i} = K$. The gating weights are then produced by applying a softmax over the logits of the selected experts:
\begin{equation}\label{eq:train_gate}
g_{\mathrm{train},i}
\;=\;
\frac{I_{\mathrm{train},i}\,\exp(s_{\mathrm{train},i})}
{\sum_{j=1}^{M} I_{\mathrm{train},j}\,\exp(s_{\mathrm{train},j})}
\quad\text{for } i=1,\dots,M.
\end{equation}
Finally, the MoE layer's output is computed as a weighted sum of the expert outputs:
\begin{equation}\label{eq:train_output}
\mathbf{y}_{\mathrm{train}}
\;=\;
\sum_{i=1}^{M} g_{\mathrm{train},i}\; \mathcal{E}_i(\mathbf{x}_{\mathrm{train}}),
\end{equation}
where $\mathcal{E}_i(\cdot)$ represents the $i$-th expert network.

Now we introduce the Rollout Routing Replay. Assume that during the inference stage, the input to the MoE layer is $\mathbf{x}_{\mathrm{infer}}$.
The router then computes the inference logits
$\mathbf{s}_{\mathrm{infer}} = \mathbf{x}_{\mathrm{infer}} \mathbf{W}_r$,
from which the routing mask
$\mathbf{I}_{\mathrm{infer}} = \operatorname{TopKMask}(\mathbf{s}_{\mathrm{infer}}, K)$
is obtained.
The main idea of Rollout Routing Replay is to reuse the inference routing mask $\mathbf{I}_{\mathrm{infer}}$ during the training forward pass
while still applying the softmax to the training logits to preserve gradient flow.
Specifically, along this replay path, the ``replay'' gating weights $g_{\mathrm{replay}}$ are computed as:
\begin{equation}\label{eq:replay_gate}
g_{\mathrm{replay},i}
\;=\;
\frac{I_{\mathrm{infer},i}\,\exp(s_{\mathrm{train},i})}
{\sum_{j=1}^{M} I_{\mathrm{infer},j}\,\exp(s_{\mathrm{train},j})}
\quad\text{for } i=1,\dots,M.
\end{equation}
These replay weights are then used to combine the training experts' outputs, producing the replay output $\mathbf{y}_{\mathrm{replay}}$:
\begin{equation}\label{eq:replay_output}
\mathbf{y}_{\mathrm{replay}}
\;=\;
\sum_{i=1}^{M} g_{\mathrm{replay},i}\; \mathcal{E}_i(\mathbf{x}_{\mathrm{train}}).
\end{equation}

This design serves two main purposes:
(a) \textbf{Aligning training and inference:} Using $\mathbf{I}_{\mathrm{infer}}$ ensures that the experts used during training replay match those selected during inference, eliminating mismatch in expert selection.
(b) \textbf{Preserving the gradient data flow:} By replaying the mask, we can precisely optimize the experts used when inferring this sequence, avoiding the optimization of unused experts, which may introduce gradient noise.

\subsection{Router Mask Caching and Multi-Turn Dialogue Support}
\label{sec:r3:cache}

Many inference engines use a KVCache prefix caching strategy~\citep{sglang,vllm} to prevent redundant prefill computations on previously seen context, which significantly reduces the total computation in multiple-turn interactions. We observe that the cached router masks share a similar property: for the same prefix tokens, the MoE router should yield identical results. Therefore, routing masks from inference engines $\mathbf{I}_{\mathrm{infer}}$ can be cached along with the prefix KVcache. Concretely, for each layer and token prefix, the corresponding routing masks are stored with KVCache. When the same prefix occurs and hits the cache, the masks can be reused, eliminating the need for recomputation. This allows Rollout Routing Replay to integrate seamlessly with prefix caching mechanisms.

Caching routing masks is especially beneficial in agent scenarios.
Many agent tasks, like software engineering~\citep{swebench} and web browsing~\citep{browsecomp}, involve multiple turns of interactions between autoregressive generation and tool calling. To improve efficiency, these processes directly reuse the KVCache from previous turns, so they do not have to regenerate data that's already been computed.
Routing mask caching enables R3 to remain efficient in RL agent tasks without re-prefilling to generate the routing masks, which is crucial for training large-scale, advanced MoE models.

In practice, partial rollout~\citep{k1.5} is commonly used to mitigate the long-tail problem of rollout, where a few overly long responses stall the entire batch.
Since the model parameters are updated while some responses are still being generated, the routing of these responses is affected as well.
We discuss several methods for combining R3 with partial rollout in Appendix~\ref{sec:partial_rollout}.

\subsection{Empirical Analysis of R3 on Training-Inference Discrepancies}

To evaluate the effectiveness of R3 in reducing training-inference discrepancies, we repeat the procedure described in Section \ref{sec:3.1} with Qwen3-30B-A3B model. In this process, we cache the routing distributions obtained during inference on SGLang and replay them within the Megatron framework. After obtaining the inference engine probability and training engine probability, using Eq.~\ref{eq:kl}, we estimate the KL divergence between training and inference. Results show that after applying R3, the KL divergence between training and inference decreases from $1.5 \times 10^{-3}$ to $7.5 \times 10^{-4}$, which is near the $6.4 \times 10^{-4}$ observed for the dense model. This intuitively indicates a reduction in the training-inference discrepancy.
We also drew the cumulative distribution plot of the ratio of training-inference discrepancies in Figure~\ref{fig:311:d} with R3. The plot indicates that, for the MoE model, applying R3 reduces the frequency of tokens with large training-inference discrepancies by a factor of 10, reaching a level comparable to the dense model.

\section{Experiments}
\label{sec:experiments}
In this section, we evaluate the performance improvement of our R3 method for reinforcement learning and compare it with other baseline methods.

\subsection{Setting}
\label{sec:experiments:setting}
\begin{table}[t]
\centering
\caption{Main evaluation results. This table presents the best scores (with corresponding global step in parentheses) achieved by various methods on evaluation benchmarks under three training configurations. The table also reports the average score (Avg) and the step at which training collapse occurred (Crash Step).}
\label{tab:result}
\footnotesize
\begin{tabular}{lcccccc}
\toprule
& \multicolumn{5}{c}{Best Metric(Best Global Step)} \\
\midrule
Method & AIME24$(\uparrow)$ & AIME25$(\uparrow)$ & AMC23$(\uparrow)$ & MATH500 Lv5$(\uparrow)$ & Avg$(\uparrow)$ & Crash Step \\
\midrule
\rowcolor{gray!30}\multicolumn{7}{c}{\textbf{Qwen3-30B-A3B-SFT, mini\_step=8}} \\
\midrule
GRPO       & 32.81(65) & 20.73(90) & 74.84(60) & 71.83(100) & 48.84(100) & 120 \\
GSPO       & 55.52(165) & \underline{38.23}(160) & \underline{90.16}(125) & 86.38(125) & 66.76(125) & - \\
GRPO+R3  & \underline{57.92}(180) & 38.02(155) & \underline{90.16}(155) & \textbf{88.62}(170) & \underline{68.05}(180) & - \\
GSPO+R3  & \textbf{58.44}(160) & \textbf{39.17}(160) & \textbf{92.50}(165) & \underline{87.87}(165) & \textbf{69.00}(165) & - \\
\midrule
\rowcolor{gray!30}\multicolumn{7}{c}{\textbf{Qwen3-30B-A3B-SFT, mini\_step=1}} \\
\midrule
GRPO         & 49.06(45) & 32.08(50) & 86.41(55) & 83.77(55) & 62.23(55) & 60 \\
GRPO+TIS     & 54.90(85) & 36.67(90) & 88.59(90) & 85.63(85) & 66.24(90) & 105 \\
GRPO+R3      & \textbf{62.92}(165) & \textbf{41.98}(170) & \textbf{93.91}(165) & \textbf{89.93}(155) & \textbf{71.83}(165) & - \\
GRPO+TIS+R3  & \underline{58.75}(180) & \underline{41.15}(180) & \underline{91.87}(175) & \underline{88.99}(180) & \underline{70.14}(175) & - \\
\midrule
\rowcolor{gray!30}\multicolumn{7}{c}{\textbf{Qwen3-30B-A3B-Base, mini\_step=1}} \\
\midrule
GRPO & 50.63(100) & 32.60(100) & 83.13(100) & 80.78(90) & 61.69(100) & 105 \\
GRPO+TIS  & \underline{56.77}(170) & {40.10}(170) & {92.50}(165) & \underline{89.37}(180) & \underline{69.22}(170) & - \\
GRPO+R3 & \textbf{60.94}(180) & \textbf{41.35}(180) & \underline{92.81}(175) & \textbf{89.74}(175) & \textbf{70.73}(180) & - \\
GRPO+TIS+R3 & {56.67}(170) & \underline{40.42}(170) & \textbf{93.12}(175) & {88.99}(165) & {69.17}(175) & - \\
\bottomrule
\end{tabular}
\end{table}

\textbf{Task and Dataset}
We choose mathematical reasoning as the target task for training. For the training dataset, we collect and filter problems from many open-source datasets, including BigMath~\citep{bigmath}, ORZ~\citep{orz}, and others, yielding approximately 100,000 verifiable math problems.
For the evaluation dataset, we adopt AIME24, AIME25, AMC23, and MATH500~\citep{hendrycksmath2021} (level 5) as our benchmark datasets. We report Avg@32 for AIME24 and AIME25, Avg@16 for AMC23, and Avg@4 for MATH500 (level 5).
During training, some models may experience performance degradation or even collapse in later stages. To ensure a fair evaluation, we measure model performance every 5 global steps throughout a single training run. We then report the highest observed performance along with the corresponding training step at which it occurs.

\textbf{Models}
We select two models for our experiments:
(a) Qwen3-30B-A3B-Base~\citep{qwen3}
(b) Qwen3-30B-A3B-SFT, which is fine-tuned from Qwen3-30B-A3B-Base on our general instruction-following dataset.

\textbf{Baseline Methods}
We consider the following baseline optimization methods for comparison: (a) \textbf{GRPO}~\citep{grpo}, additionally applies the \textit{Clip Higher} technique from DAPO~\citep{yu2025dapo}, with parameters $\epsilon_{\text{low}} = 0.2$ and $\epsilon_{\text{high}} = 0.27$; (b) \textbf{TIS}~\citep{tis}, employs rollout importance sampling and uses an upper clipping threshold $C = 2$; (c) \textbf{GSPO}~\citep{gspo}, employs sequence-level importance sampling, with parameters $\epsilon_{\text{low}} = 3\times10^{-4}$ and $\epsilon_{\text{high}} = 4\times10^{-4}$. Since our R3 method is orthogonal to policy optimization methods such as GSPO or TIS, we also evaluate various combinations of these techniques.

\textbf{Multiple Mini Steps vs. Single Mini Step}
In PPO-like algorithms, a batch of samples collected in one global step is typically divided into multiple mini steps. However, previous research~\citep{opo} has shown that applying a strictly on-policy strategy with only a single mini step can yield better performance.
We investigate both settings. For multiple mini steps, each PPO mini step trains $2048/8 = 256$ samples with 8 optimizer updates and a learning rate of $1\times 10^{-6}$. In the multiple mini-step scenario, for R3 method we replayed the routing both when recomputing the old policy and updating policy.
For the single mini step scenario, all 2048 samples are updated once. Due to fewer updates, we increase the learning rate to $3\times 10^{-6}$. We don't recompute old policy in this scenario.

\textbf{Other Setting}
We implement R3 using the VeRL~\citep{verl} framework, use Megatron~\citep{megatron} for training, and SGLang~\citep{sglang} for inference.
We set the batch size to 256 and $n=8$, totaling 2048 samples per round. The maximum prompt length is 2048, and the maximum generation length is 30720. The max global step is 180. We adopt Dynamic Sampling~\citep{yu2025dapo}, retaining only partially correct samples during generation until a sufficient batch size is accumulated for training.
No auxiliary loss for expert balancing is introduced.

\subsection{Experimental Results}

The main evaluation results are shown in Table \ref{tab:result}.
More detailed evaluation results and training logs, including per-evaluation scores as well as training metrics such as entropy, average reward, gradient norm, and training-inference KL divergence, are provided in the appendix \ref{sec:appA}.
\begin{figure}[t]
\centering
\small
\includegraphics[width=\textwidth]{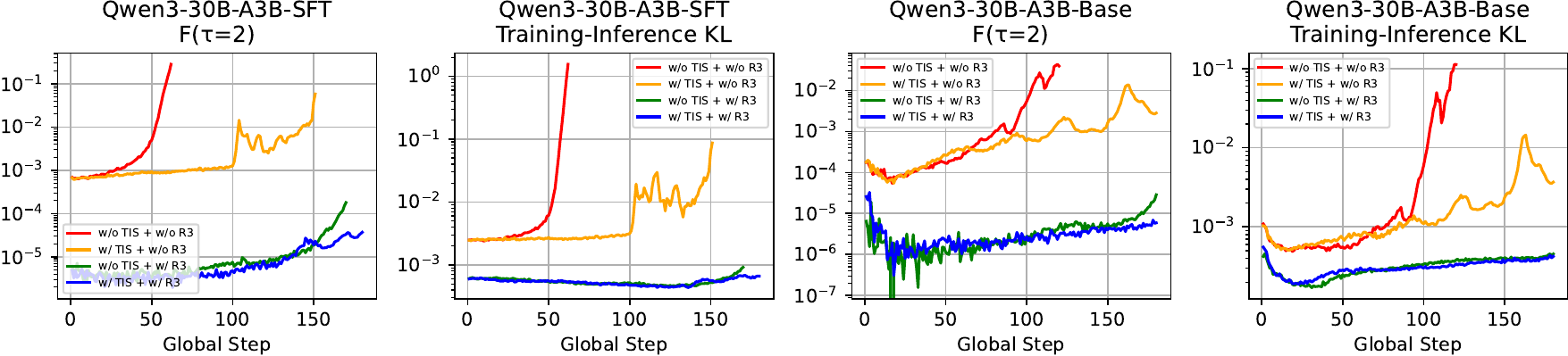}
\caption{Analysis of training–inference collapse. The plot shows the estimated training–inference KL divergence and the extreme token distribution function $F(\tau=2)$ (Eq. \ref{eq:etdf}) at each training step.}
\label{fig:analy_2}
\vspace{10pt}
\includegraphics[width=\textwidth]{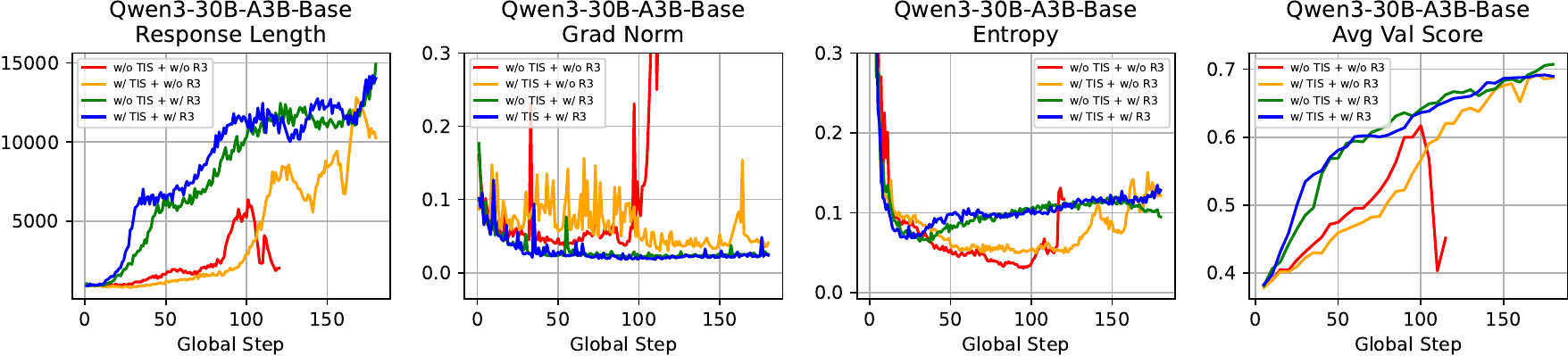}
\caption{Training dynamics of Qwen3-30B-A3B-Base, including response length, gradient norm, entropy, and average validation score throughout the training process}
\label{fig:analy_1}
\end{figure}

\textbf{Overall Performance.}
R3 achieves better results across different scenarios. In the multi mini-step setting, GRPO+R3 outperforms GSPO by 1.29. Furthermore, combining R3 with GSPO further improves performance by 0.95 points. In the single mini-step setting, R3 outperforms TIS by 5.58 on SFT model and 1.51 on base model. However, combining R3 with TIS does not yield clear gains and may even degrade performance; for instance, in the single mini-step on SFT model, TIS+R3 scores 3.29 points lower than R3 alone. Since R3 already significantly reduces the policy discrepancy between training and inference, the additional correction from TIS offers negligible benefit~\citep{tis}.

\textbf{Training Stability.}
In the single mini-step setting, three reinforcement learning processes without R3 collapsed during training. To investigate the causes of these collapses, we plotted the estimated training–inference KL divergence and the extreme token distribution function $F(\tau=2)$ (Eq.~\ref{eq:etdf}) at each training step (see Figure~\ref{fig:analy_2}). We observed that, in each training process, both the KL divergence and $F(\tau=2)$ values increased over training process. Moreover, collapsed training runs were almost always accompanied by abnormally high KL and $F(\tau=2)$ values.
For instance, in the SFT model trained using GRPO under the single mini-step setting, after 60 global steps, the value of $F(\tau=2)$ exceeded 0.1. This indicates that for 10\% of the tokens, the probability under the training framework differed from that under the inference framework by more than a factor of 2, demonstrating severe training–inference inconsistency. In contrast, for training processes using R3, the value of $F(\tau=2)$ remained below $10^{-4}$. By aligning the training and inference distributions, R3 effectively stabilized reinforcement learning for MoE models.

\textbf{Optimization and Generation Behavior.}
During training, R3 also enhances optimization stability, exploration behavior, and generation dynamics. We plotted the sequence length, gradient norm, generation entropy, and evaluation score (see Figure~\ref{fig:analy_1}).  It shows that R3 has smaller gradient norms, a smoother sequence growth pattern, and more stable entropy.
(a) \textbf{Sequence Length:} With R3, the generated sequence length rises rapidly at the beginning of training, indicating that R3 can quickly capture the correct optimization direction. By contrast, the other two training processes increase only slowly after step 80 and show more pronounced fluctuations.
(b) \textbf{Gradient Norm:} R3 maintains consistently stable gradient norms, indicating a more stable optimization process.
(c) \textbf{Generation Entropy:} With R3, entropy begins to increase steadily after about 25 step, suggesting that the model starts exploring better strategies earlier. Without R3, entropy increases much later and fluctuates heavily.

\subsection{Generalization}
\subsubsection{Generalization to Multi-Turn Task}
\begin{table}[h]
\centering
\small
\caption{Evaluation results of RL training on SWE Task}
\begin{tabular}{lcc}
\toprule
\rowcolor{gray!30}\multicolumn{3}{c}{\textbf{Model: Qwen3-30B-A3B}} \\
\midrule
Method & SWE-bench Verified$(\uparrow)$ & Crash Step \\
\midrule
GRPO     & 31.80 & 90 \\
GRPO+R3  & \textbf{38.60} & - \\
\bottomrule
\end{tabular}
\label{table:swe_res_main}
\end{table}
To verify that R3 can be adapted to multi-turn tasks, we conduct reinforcement learning experiments on the software engineering task.
This task has recently become a key benchmark for state-of-the-art models, as it simultaneously evaluates a model's code generation ability, tool-use capability, and multi-turn reasoning performance.
We use the Qwen3-30B-A3B~\citep{qwen3} model, use R2E-Gym-Lite~\citep{jain2025r2e} as the training dataset and SWE-bench-Verified~\citep{swebench} as the test dataset. Detailed experimental settings are provided in Appendix~\ref{sec:appdD}.
The results are presented in Table \ref{table:swe_res_main}. The training process of GRPO without R3 collapses after around 90 training steps, while GRPO+R3 maintains stability and achieves higher scores. This result confirms that R3 generalizes well to multi-turn reinforcement learning settings.

\subsubsection{Generalization to Other MoE Model}
\begin{table}[h]
\centering
\small
\caption{Evaluation results of the RL training process of DeepSeek-V2-Lite-ReasoningSFT}
\label{table:dpsk_brief}
\begin{tabular}{lcc}
\toprule
\rowcolor{gray!30}\multicolumn{3}{c}{\textbf{Model: DeepSeek-V2-Lite-ReasoningSFT}} \\
\midrule
Method & Math Avg$(\uparrow)$ & Crash Step \\
\midrule
GRPO &  23.55(55) & 100 \\
GRPO+R3 & \textbf{26.43}(235) & - \\
\bottomrule
\end{tabular}
\end{table}

To evaluate the performance of R3 on other MoE models, we conduct RL training experiments on DeepSeek-V2-Lite \citep{deepseekai2024deepseekv2strongeconomicalefficient}. We first fine-tune DeepSeek-V2-Lite on the open-source long reasoning dataset {Mixture-of-Thoughts}, producing the DeepSeek-V2-Lite-ReasoningSFT model. We then conduct reinforcement learning experiments on this model.
Detailed experimental settings are provided in Appendix~\ref{sec:dpskv2lite}. The results show that R3 also stabilizes training and achieves higher scores, indicating that R3 generalizes well across different model architectures.

\subsection{Speed and Memory Analysis}
\subsubsection{End-to-End Throughput Benchmark}
\label{sec:end2end_speed}

To show that R3 can be practically deployed in large-scale training pipelines without incurring prohibitive computational costs, we conduct an end-to-end throughput benchmark to quantify the runtime overhead introduced by R3.

Experiments are performed on the Qwen3-30B-A3B model under two configurations: with and without R3 enabled. Both settings employ identical hardware and software environments to ensure a fair comparison. We set batch size = 256, $n=8$, and a maximum generation length of 32K.
We capture the complete end-to-end execution latency per training step, encompassing inference, forward computation, backward propagation, optimizer update, and all communication overhead between inference and training components.

\begin{table}[h]
\centering
\small
\caption{Throughput analysis result of Rollout Routing Replay}
\begin{tabular}{lcc}
\toprule
Metric & w/o R3 & w/ R3 \\
\midrule
\# Global Step & 10 & 10 \\
\# Token & 203,989,395 & 203,201,992 \\
Runtime (sec) & 4820 & 4967 \\
\# Token/sec & 42,321 & 40,910 \\
Relative Time Cost & 100\% & 103.45\% \\
\bottomrule
\end{tabular}
\label{tab:end2end_benchmark}
\end{table}

As summarized in Table~\ref{tab:end2end_benchmark}, enabling R3 results in only a \textbf{3.45\%} reduction in end-to-end training throughput. This demonstrates that R3 can be efficiently integrated into large-scale reinforcement learning pipelines with negligible impact on training efficiency.

\subsubsection{Memory Usage Analysis}
R3 only introduces a small additional memory overhead on CPU memory.
We keep the router mask in CPU memory most of the time, and only load it onto the GPU when the corresponding sequence is loaded into the training framework. Therefore, storing the router mask only consumes CPU memory.
In the Appendix~\ref{sec:appdG}, we provide a detailed analysis of the memory overhead of training caused by R3. In summary, in our experimental setup, each node consumed an additional 2.5GB of CPU memory during the inference phase and 1.5GB during the training phase.

\section{Related Works}

\subsection{Mixture of Experts}
The Mixture-of-Experts (MoE) architecture, originating as an ensemble method to route inputs to specialized subnetworks~\citep{jacobs1991adaptive,shazeer2017outrageouslylargeneuralnetworks}, has become a cornerstone for scaling modern language models. By employing a gating network to sparsely activate only a subset of expert parameters per token, MoE decouples a model's total parameter count from its inference cost, enabling a substantial increase in model capacity. This computational efficiency has driven its adoption in state-of-the-art Transformer models~\citep{jiang2024mixtral,liu2024deepseek,k1.5,qwen3}.

\subsection{Reinforcement Learning with Verifiable Rewards}
Reinforcement Learning with verifiable rewards (RLVR) has become a standard method for refining the complex reasoning~\citep{xie2025logic}, mathematical~\citep{grpo,orz}, and code abilities~\citep{deepcoder2025} of large language models. Algorithms such as GRPO~\citep{grpo}, DAPO~\citep{yu2025dapo}, and GSPO~\citep{gspo} are now widely adopted.
These methods have already been used to train state-of-the-art reasoning models~\citep{qwen3,mimo,chen2025minimax,dpskr1,k1.5}.
\subsection{Nondeterminism, Framework Discrepancies and Reinforcement Learning}
With the rise of LLM RL, training-inference discrepancies and non-determinism issues have also drawn widespread attention, largely due to concerns about training stability.
\citet{chen2025minimax} pointed out the discrepancies problem between training and inference can cause instability problems in reinforcement learning and alleviated it by raising the output layer's precision to FP32.
\citet{yuan2025understanding} examined how numerical precision affects inference nondeterminism.
\citet{nondeter} discussed the main reasons why inference can become nondeterministic.
From the perspective of importance sampling, \citet{tis}, \citet{IcePop2025}, and \citet{liu-li-2025-rl-collapse} proposed the TIS, Icepop, and MIS methods to address the problem. \citet{qi2025defeatingtraininginferencemismatchfp16} pointed out that using FP16 can resolve the training-inference discrepancies issue.

\subsection{Routing Replay}

Routing Replay was first introduced in~\citet{gspo} (hereafter referred to as \textbf{Recompute Routing Replay}). We clarify how our \textbf{Rollout Routing Replay} differs from it. As shown in Figure~\ref{fig:teaser}, there are three forward-pass stages in a typical reinforcement learning framework: \textit{Rollout}, \textit{Recompute} and \textit{Update}.
Recompute Routing Replay caches the routing from the \textit{Recompute} stage and replays it in the \textit{Update} stage, addressing the issue of routing discrepancies caused by model updates. However, this approach does not take into account that inconsistencies between training and inference frameworks can also lead to routing discrepancies. In contrast, our Rollout Routing Replay caches the routing from the \textit{Rollout} stage and replays it in both the \textit{Recompute} and \textit{Update} stages, thereby resolving routing discrepancies arising from framework difference. Notably, since our method uses the same routing for both the Recompute and Update stages, it also addresses routing discrepancies caused by model updates.
Furthermore, when the mini step is set to 1, the current policy and the old policy become identical, which makes the Recompute Routing Replay method ineffective. Since discrepancies between training and inference frameworks continue to exist, the Rollout Routing Replay method remains effective under this setting.

\section{Conclusion}
In this work, we identify training-inference routing discrepancies as the primary source of instability in MoE reinforcement learning. To address this, we propose Rollout Routing Replay (R3), which reuses inference-time routing distributions during training to align expert selection while preserving gradient flow. Experiments across multiple RL settings demonstrate that R3 substantially reduces training-inference divergence, stabilizes training, and consistently outperforms existing methods.
Our results demonstrate the importance of aligning training and inference in MoE models and show that R3 provides a practical solution for improving stability.

\bibliography{main}

\newpage
\appendix

\section{Detailed Evaluation Results and Training Metrics}
\label{sec:appA}
\begin{figure}[h]
\centering
\includegraphics[width=\textwidth]{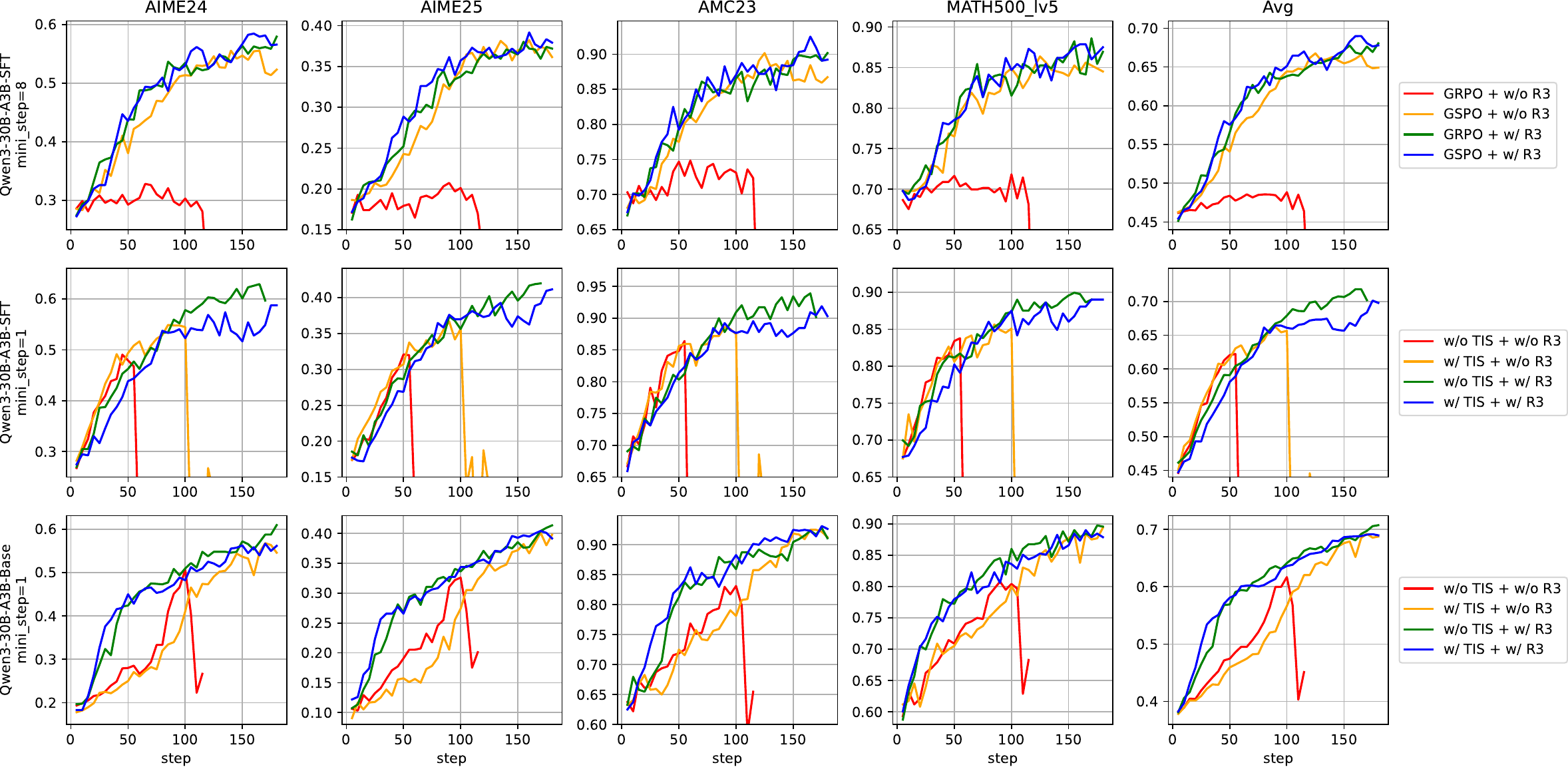}
\caption{The detailed evaluation results of the experiment in Section \ref{sec:experiments}.}
\label{fig:appendixA}
\end{figure}

\label{sec:appB}
\begin{figure}[h]
\centering
\includegraphics[width=\textwidth]{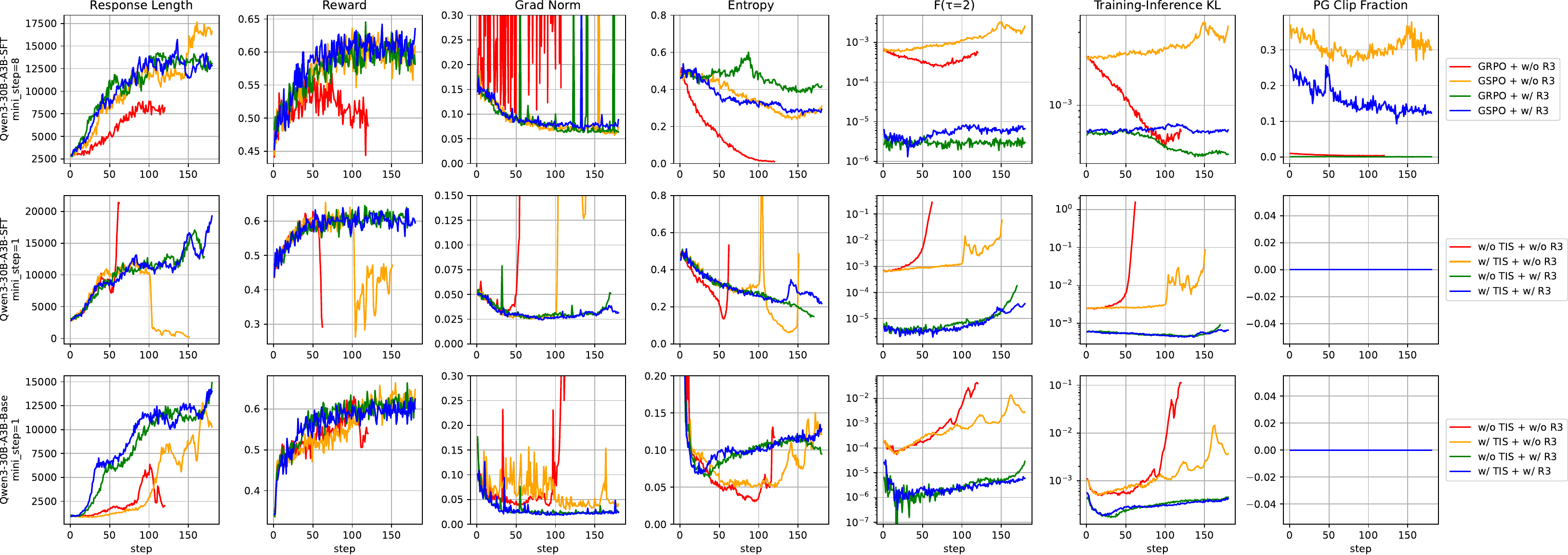}
\caption{The detailed training metrics of the experiment in Section \ref{sec:experiments}.}
\label{fig:appendixB}
\end{figure}

\newpage
\section{Combining R3 with Partial Rollout}
\label{sec:partial_rollout}

Partial rollout~\citep{k1.5} mitigates the long-tail problem of rollout by allowing a trajectory to span multiple parameter versions: a trajectory that is unfinished when the model parameters are updated is resumed with the new parameters.
Because the inference engine typically invalidates the KV cache after a parameter update, the context of the paused trajectory is re-prefilled with the new parameters, which also produces new routing masks.
In this section, we discuss several methods for combining R3 with partial rollout and analyze their correctness and drawbacks.

\begin{figure}[h]
\centering
\includegraphics{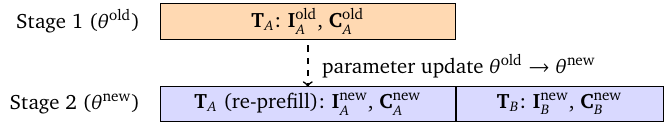}
\caption{Default setting of partial rollout. Orange and blue denote states computed with $\theta^{\mathrm{old}}$ and $\theta^{\mathrm{new}}$, respectively. After the parameter update, $\mathbf{T}_A$ is re-prefilled with $\theta^{\mathrm{new}}$, and the inference engine continues the rollout to obtain $\mathbf{T}_B$.}
\label{fig:partial_rollout}
\end{figure}

\textbf{Setting}
Figure~\ref{fig:partial_rollout} illustrates the setting.
We split a trajectory at the parameter update into two segments, $A$ and $B$, which are rolled out before and after the update, respectively.
For each segment, we denote its token sequence, routing masks, and KV cache by $\mathbf{T}$, $\mathbf{I}$, and $\mathbf{C}$, respectively, with the segment as the subscript; the superscript of $\mathbf{I}$ and $\mathbf{C}$ indicates whether they are computed with the parameters before the update ($\theta^{\mathrm{old}}$) or after it ($\theta^{\mathrm{new}}$).
We use $\oplus$ for concatenation along the sequence, so the complete trajectory is $\mathbf{T}_A \oplus \mathbf{T}_B$.
In the first stage, the inference engine processes $\mathbf{T}_A$ with $\theta^{\mathrm{old}}$, producing $\mathbf{I}_A^{\mathrm{old}}$ and $\mathbf{C}_A^{\mathrm{old}}$.
By default, after the parameter update, the inference engine re-prefills $\mathbf{T}_A$ with $\theta^{\mathrm{new}}$, producing $\mathbf{I}_A^{\mathrm{new}}$ and $\mathbf{C}_A^{\mathrm{new}}$, and then continues the rollout on top of $\mathbf{C}_A^{\mathrm{new}}$, producing $\mathbf{T}_B$ together with $\mathbf{I}_B^{\mathrm{new}}$ and $\mathbf{C}_B^{\mathrm{new}}$.
Since even floating-point level perturbations can flip the selected experts (Section~\ref{sec:3.2}), the parameter update generally makes $\mathbf{I}_A^{\mathrm{new}}$ differ from $\mathbf{I}_A^{\mathrm{old}}$.
For clarity, we consider a single parameter update; the methods below naturally extend to multiple updates.

\textbf{Routing Consistency}
Following Section~\ref{sec:r3}, we call a method routing-consistent if, for every token that contributes to the loss, the training forward pass replays the routing masks under which the inference engine generated it, at both the token's own position and all preceding positions.
The preceding positions matter because their routing determines the KV cache that later tokens attend to.
In the current setting, this requirement amounts to two conditions.
(a) For the tokens in $\mathbf{T}_A$, the training sequence on which their loss is computed must replay $\mathbf{I}_A^{\mathrm{old}}$ on $\mathbf{T}_A$.
(b) For the tokens in $\mathbf{T}_B$, the training sequence on which their loss is computed must replay $\mathbf{I}_B^{\mathrm{new}}$ on $\mathbf{T}_B$ and, on $\mathbf{T}_A$, the routing masks under which the KV cache attended by $\mathbf{T}_B$ was computed.
Under the default re-prefill, this KV cache is $\mathbf{C}_A^{\mathrm{new}}$ computed under $\mathbf{I}_A^{\mathrm{new}}$, so (a) and (b) require different routing masks on $\mathbf{T}_A$ and cannot both hold within a single training sequence.
We discuss five methods below: the first two are simple but cannot maintain routing consistency, while the last three maintain routing consistency.
Table~\ref{tab:partial_rollout} summarizes these methods.

\subsection{Inconsistent Methods}

The following methods keep the default re-prefill and train a single sequence $\mathbf{T}_A \oplus \mathbf{T}_B$, so each of them violates one of the two conditions and is not routing-consistent.

\textbf{Re-prefill Only}
This method replays only the routing masks of the second stage, namely $\mathbf{I}_A^{\mathrm{new}} \oplus \mathbf{I}_B^{\mathrm{new}}$ on $\mathbf{T}_A \oplus \mathbf{T}_B$.
It is simple and keeps the tokens in $\mathbf{T}_B$ consistent, but the tokens in $\mathbf{T}_A$ were generated under $\mathbf{I}_A^{\mathrm{old}}$ and are trained under $\mathbf{I}_A^{\mathrm{new}}$, which violates condition (a).

\textbf{Concat Old}
This method concatenates the old routing masks of $\mathbf{T}_A$ with the new routing masks of $\mathbf{T}_B$, namely $\mathbf{I}_A^{\mathrm{old}} \oplus \mathbf{I}_B^{\mathrm{new}}$ on $\mathbf{T}_A \oplus \mathbf{T}_B$.
It is equally simple and keeps the tokens in $\mathbf{T}_A$ consistent, but the tokens in $\mathbf{T}_B$ were generated on top of $\mathbf{C}_A^{\mathrm{new}}$ computed under $\mathbf{I}_A^{\mathrm{new}}$, whereas training recomputes their context under $\mathbf{I}_A^{\mathrm{old}}$, which violates condition (b).

\subsection{Consistent Methods}

The following methods satisfy both conditions, either by training $\mathbf{T}_A$ and $\mathbf{T}_B$ in separate sequences or by ensuring that the KV cache attended by $\mathbf{T}_B$ is computed under $\mathbf{I}_A^{\mathrm{old}}$, and therefore guarantee routing consistency between training and inference.

\textbf{Split Training}
This method keeps the default re-prefill and trains two sequences that share the advantage of the complete trajectory: $\mathbf{T}_A$ replaying $\mathbf{I}_A^{\mathrm{old}}$ with the loss on $\mathbf{T}_A$, and $\mathbf{T}_A \oplus \mathbf{T}_B$ replaying $\mathbf{I}_A^{\mathrm{new}} \oplus \mathbf{I}_B^{\mathrm{new}}$ with the loss on $\mathbf{T}_B$.
It requires no change to the inference engine, but $\mathbf{T}_A$ goes through the forward and backward passes twice, and this extra training computation grows with the number of parameter updates that a trajectory spans.

\textbf{Replay in Re-prefill}
This method replays $\mathbf{I}_A^{\mathrm{old}}$ in the inference engine when re-prefilling $\mathbf{T}_A$, so that $\mathbf{C}_A^{\mathrm{new}}$ is computed under $\mathbf{I}_A^{\mathrm{old}}$, and training then replays $\mathbf{I}_A^{\mathrm{old}} \oplus \mathbf{I}_B^{\mathrm{new}}$ on $\mathbf{T}_A \oplus \mathbf{T}_B$.
It needs no extra training computation, but its implementation is complex: the MoE layers of the inference engine must accept externally provided expert selections during prefill, and $\mathbf{I}_A^{\mathrm{old}}$ must be kept until the trajectory is resumed.

\textbf{Cache Retention}
This method skips re-prefill and continues the rollout on top of the retained $\mathbf{C}_A^{\mathrm{old}}$, whose routing masks $\mathbf{I}_A^{\mathrm{old}}$ are cached alongside it (Section~\ref{sec:r3:cache}), and training then replays $\mathbf{I}_A^{\mathrm{old}} \oplus \mathbf{I}_B^{\mathrm{new}}$ on $\mathbf{T}_A \oplus \mathbf{T}_B$.
It is simple and saves the computation of re-prefill, but the KV cache is stale: the tokens in $\mathbf{T}_B$ are generated by $\theta^{\mathrm{new}}$ attending to $\mathbf{C}_A^{\mathrm{old}}$ computed by $\theta^{\mathrm{old}}$, which may degrade the quality of the newly generated sequence.

\begin{table}[h]
\centering
\small
\caption{Methods for combining R3 with partial rollout. Parentheses give the routing masks replayed on each training sequence, and \ding{51}/\ding{55} indicate whether the tokens in $\mathbf{T}_A$ and $\mathbf{T}_B$ are routing-consistent.}
\label{tab:partial_rollout}
\setlength{\tabcolsep}{3pt}
\begin{tabularx}{\textwidth}{lllcc>{\raggedright\arraybackslash}X}
\toprule
Method & \makecell[l]{Context of $\mathbf{T}_B$\\at Inference} & \makecell[l]{Training Sequences\\(Replayed Masks)} & $\mathbf{T}_A$ & $\mathbf{T}_B$ & Drawback \\
\midrule
Re-prefill only & $\mathbf{C}_A^{\mathrm{new}}$ under $\mathbf{I}_A^{\mathrm{new}}$ & $\mathbf{T}_A \oplus \mathbf{T}_B$ ($\mathbf{I}_A^{\mathrm{new}} \oplus \mathbf{I}_B^{\mathrm{new}}$) & \ding{55} & \ding{51} & Inconsistent on $\mathbf{T}_A$ \\
Concat old & $\mathbf{C}_A^{\mathrm{new}}$ under $\mathbf{I}_A^{\mathrm{new}}$ & $\mathbf{T}_A \oplus \mathbf{T}_B$ ($\mathbf{I}_A^{\mathrm{old}} \oplus \mathbf{I}_B^{\mathrm{new}}$) & \ding{51} & \ding{55} & Inconsistent on $\mathbf{T}_B$ \\
\midrule
Split training & $\mathbf{C}_A^{\mathrm{new}}$ under $\mathbf{I}_A^{\mathrm{new}}$ & \makecell[l]{$\mathbf{T}_A$ ($\mathbf{I}_A^{\mathrm{old}}$),\\$\mathbf{T}_A \oplus \mathbf{T}_B$ ($\mathbf{I}_A^{\mathrm{new}} \oplus \mathbf{I}_B^{\mathrm{new}}$)} & \ding{51} & \ding{51} & Extra training computation \\
Replay in re-prefill & $\mathbf{C}_A^{\mathrm{new}}$ under $\mathbf{I}_A^{\mathrm{old}}$ & $\mathbf{T}_A \oplus \mathbf{T}_B$ ($\mathbf{I}_A^{\mathrm{old}} \oplus \mathbf{I}_B^{\mathrm{new}}$) & \ding{51} & \ding{51} & Complex implementation \\
Cache retention & $\mathbf{C}_A^{\mathrm{old}}$ under $\mathbf{I}_A^{\mathrm{old}}$ & $\mathbf{T}_A \oplus \mathbf{T}_B$ ($\mathbf{I}_A^{\mathrm{old}} \oplus \mathbf{I}_B^{\mathrm{new}}$) & \ding{51} & \ding{51} & Stale KV cache \\
\bottomrule
\end{tabularx}
\end{table}

\section{Other Model Experiments}
\label{sec:appdB}

\subsection{Experiments on DeepSeek-V2-Lite}
\label{sec:dpskv2lite}
To evaluate the performance of R3 on other MoE models, we conduct experiments on DeepSeek-V2-Lite \citep{deepseekai2024deepseekv2strongeconomicalefficient}. We first fine-tune DeepSeek-V2-Lite on the open-source long reasoning dataset {Mixture-of-Thoughts} \footnote{https://huggingface.co/datasets/open-r1/Mixture-of-Thoughts}\citep{openr1, penedo2025codeforces, lozhkov2025openr1math220k, bercovich2025llamanemotronefficientreasoningmodels}, producing the DeepSeek-V2-Lite-ReasoningSFT model. We then conduct reinforcement learning experiments on this model.

We compare GRPO and GRPO+R3 with a learning rate of $3\times 10^{-6}$ when $\text{mini\_step}=1$; all other hyperparameters follow the configuration described in Section~\ref{sec:experiments:setting}.

The results are presented in Table \ref{table:dpsk}. The experiments show that training runs without R3 eventually collapse (for example, when $\text{mini\_step}=1$, GRPO collapses at training step $100$), whereas training runs with R3 remain stable for up to $250$ training steps. These findings demonstrate that the collapse issue occurs not only in Qwen3 MoE but also in other MoE models, and that R3 can likewise ensure stable training across these settings.

\begin{table}[h]
\centering
\small
\caption{Evaluation results of the RL training process of DeepSeek-V2-Lite-ReasoningSFT}
\label{table:dpsk}
\begin{tabular}{lcccccc}
\toprule
& \multicolumn{5}{c}{Best Metric(Best Global Step)} \\
\midrule
Method & AIME24$(\uparrow)$ & AIME25$(\uparrow)$ & AMC23$(\uparrow)$ & MATH500 Lv5$(\uparrow)$ & Avg$(\uparrow)$ & Crash Step \\
\midrule
\rowcolor{gray!30}\multicolumn{7}{c}{\textbf{DeepSeek-V2-Lite-ReasoningSFT, mini\_step=1, max\_global\_step=250}} \\
\midrule
GRPO & 6.87(60) & 9.69(55) &42.50(50) & 37.50(55) & 23.55(55) & 100 \\
GRPO+R3 & \textbf{8.85}(210) & \textbf{11.46}(145) & \textbf{48.28}(165) & \textbf{43.28}(235) & \textbf{26.43}(235) & - \\
\bottomrule
\end{tabular}
\end{table}

\subsection{Training-Inference Discrepancy Analysis on Additional Models}

We evaluate $F(\tau=2)$ and the KL divergence for several models with and without R3, including Qwen3-30B-A3B, DeepSeek-V2-Lite, Mixtral-8x7B, and Moonlight-16B-A3B. The results are as follows. ``SFT'' indicates that the model has been fine-tuned.

\begin{table}[h]
\centering
\small
\caption{Training-Inference Discrepancy}
\begin{tabular}{lcccccc}
    \toprule
    Model \& Method & $F(\tau=2)$ & $KL$  \\
    \midrule
    Qwen3-30B-A3B                   & 2.54e-4             & 1.37e-3           \\
    Qwen3-30B-A3B + R3              & \textbf{5.83e-6}             & \textbf{7.03e-4}           \\
    \midrule
    DeepSeek-V2-Lite(SFT)                & 2.25e-3             & 4.06e-3 \\
    DeepSeek-V2-Lite(SFT) + R3           & \textbf{2.16e-4}             & \textbf{1.17e-3} \\
    \midrule
    Mixtral-8x7B(SFT)                    & 9.77e-5             & 1.03e-3 \\
    Mixtral-8x7B(SFT) + R3               & \textbf{1.08e-6}             & \textbf{4.98e-4} \\
    \midrule
    Moonlight-16B-A3B-Instruct      & 1.73e-5         & 3.67e-4 \\
    Moonlight-16B-A3B-Instruct + R3 & \textbf{6.94e-7}         & \textbf{1.92e-4} \\
    \bottomrule
\end{tabular}
\label{table:appendixb3}
\end{table}

From the results, it can be seen that R3 can improve training and inference consistency in many types of MoE models. This includes models with different router activations (Qwen3-30B-A3B, DeepSeek-V2-Lite, and Mixtral-8x7B use softmax, while Moonlight-16B-A3B uses sigmoid), different numbers of experts (DeepSeek-V2-Lite and Moonlight-16B-A3B have 64 experts, Qwen3-30B-A3B has 128 experts, and Mixtral has only 8 experts), and whether shared experts are used (Moonlight-16B-A3B and DeepSeek-V2-Lite use shared experts). This shows the general applicability of the R3 method.

\section{Fix Router Experiments}
\label{sec:appdC}

We repeated a controlled study on the Qwen3-30B-A3B-Base model to investigate whether freezing the router alone prevents training collapse. In this experiment the router parameters were frozen, while all other training hyperparameters and procedures strictly followed the configuration given in Section~\ref{sec:experiments:setting}.
The results are shown in Table \ref{table:fixrouter}

We observe that freezing the router does not prevent catastrophic collapse: the model still experiences a pronounced deterioration in performance at approximately $\text{training\_step}=120$.
The experiment implies that attributing collapse solely to failures in the router parameters is insufficient. Instead, it suggests that the collapse phenomenon involves degradation of the model parameters as a whole rather than an isolated malfunction of the router.

\begin{table}[h]
\centering
\small
\setlength{\tabcolsep}{4pt}
\caption{Evaluation results of the RL training process of fixed router}
\begin{tabular}{lcccccc}
\toprule
& \multicolumn{5}{c}{Best Metric(Best Global Step)} \\
\midrule
Method & AIME24$(\uparrow)$ & AIME25$(\uparrow)$ & AMC23$(\uparrow)$ & MATH500 Lv5$(\uparrow)$ & Avg$(\uparrow)$ & Crash Step \\
\midrule
\rowcolor{gray!30}\multicolumn{7}{c}{\textbf{Qwen3-30B-A3B-Base, mini\_step=1, max\_global\_step=180}} \\
\midrule
GRPO+FixRouter & 49.27(115) & 33.75(115) & 84.38(115) & 82.46(115) & 62.47(115) & 120 \\
GRPO+R3 & \textbf{60.94}(180) & \textbf{41.35}(180) & \textbf{92.81}(175) & \textbf{89.74}(175) & \textbf{70.73}(180) & - \\
\bottomrule
\end{tabular}
\label{table:fixrouter}
\end{table}

\newpage
\section{Multi-Turn Task Reinforcement Learning Experiments}
\label{sec:appdD}

In Section~\ref{sec:r3:cache}, we mentioned that the R3 method can be adapted to Multi-Turn tasks. To verify this capability, we conduct reinforcement learning experiments on the software engineering~\citep{swebench} task using the Qwen3-30B-A3B~\citep{qwen3} model. We use R2E-Gym-Lite\footnote{https://huggingface.co/datasets/R2E-Gym/R2E-Gym-Lite}~\citep{jain2025r2e} as the training dataset and SWE-bench Verified\footnote{https://huggingface.co/datasets/princeton-nlp/SWE-bench\_Verified}~\citep{swebench} as the validation dataset. For training,
we set the maximum sequence length to $65536$ tokens (including the prompt, model response, and environment observation), the batch size to $64$, the maximum number of interaction steps to $50$, the learning rate to $2\times10^{-6}$, and the $\text{mini\_step}$ to $1$. For evaluation, we report Pass@1 on SWE-bench Verified. We evaluate the model performance every 10 global steps. We compare the performance of GRPO and GRPO+R3 under these settings. All other hyperparameters follow the configuration described in Section~\ref{sec:experiments:setting}.

\begin{figure}[h]
\centering
\includegraphics[width=0.85\textwidth]{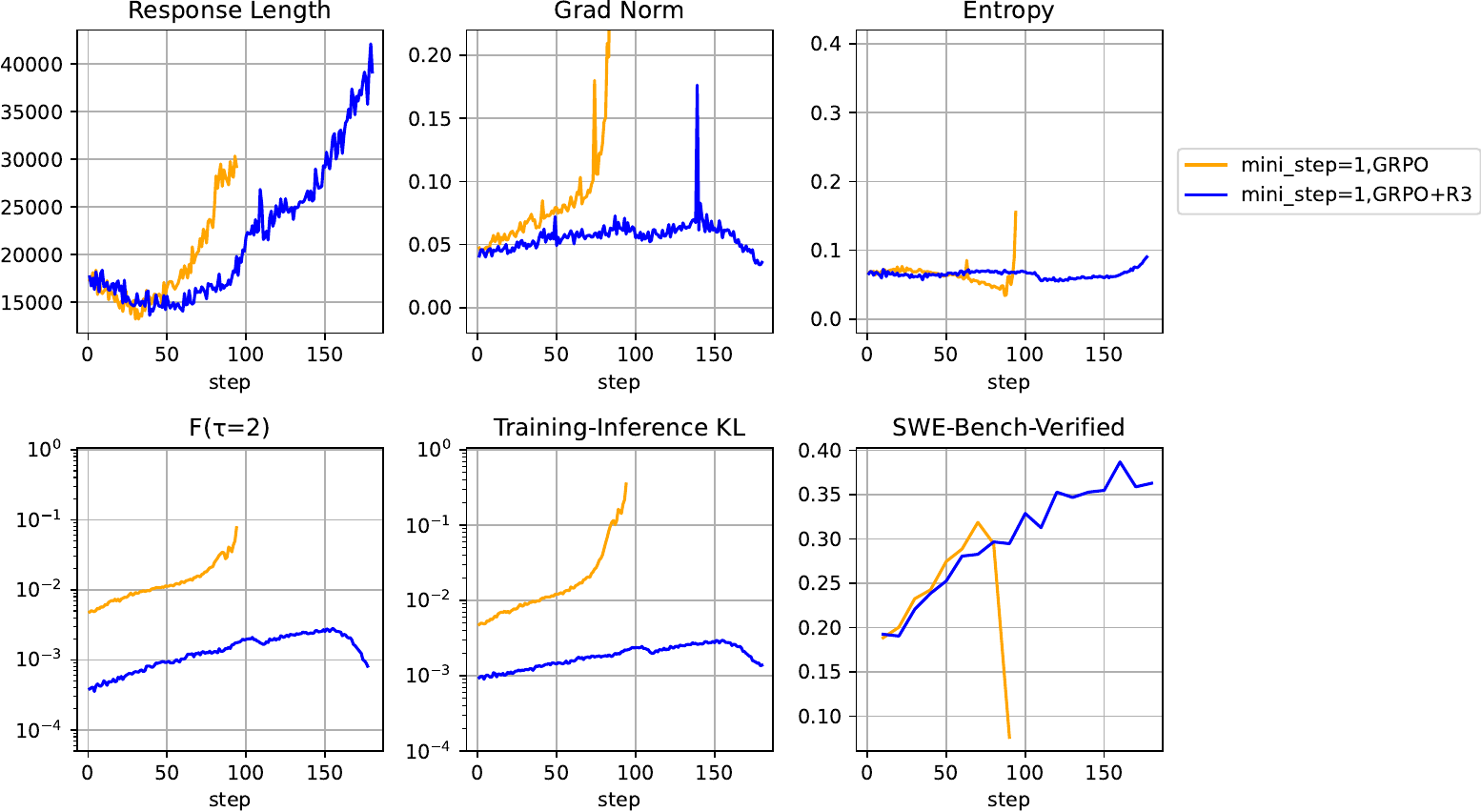}
\caption{The detailed training and evaluation results of the SWE experiment.}
\label{fig:appendixD}
\end{figure}

The results are presented in Table \ref{table:swe_res}, and the detailed training metrics are provided in Figure \ref{fig:appendixD}. The training process of GRPO without R3 collapses after around 90 training steps, while GRPO+R3 remains stable. The final performance of GRPO+R3 reaches a Pass@1 of 38.6, which is 6.8 points higher than that of GRPO. Since we use the Router Mask Caching technique described in Section~\ref{sec:r3:cache}, we do not need to re-prefill the prefix to obtain the routing mask, thus the rollout speed is not slowed down. These results confirm that R3 generalizes well to multi-turn reinforcement learning settings.
\begin{table}[h]
\centering
\small
\caption{Evaluation results of RL training on SWE Task with Qwen3-30B-A3B}
\begin{tabular}{lccc}
\toprule
\rowcolor{gray!30}\multicolumn{4}{c}{\textbf{Qwen3-30B-A3B, mini\_step=1, max\_global\_step=180}} \\
\midrule
Method & SWE-bench Verified$(\uparrow)$ & Best Global Step & Crash Step \\
\midrule
GRPO     & 31.80 & 70 & 90 \\
GRPO+R3  & \textbf{38.60} & 160 & - \\
\bottomrule
\end{tabular}
\label{table:swe_res}
\end{table}

\newpage
\section{Additional Experiments of Reasoning SFT Model}
\label{sec:appC}

To evaluate the performance of R3 on reasoning SFT models, we fine-tune Qwen3-30B-A3B-Base~\citep{qwen3} on the open-source long reasoning dataset {Mixture-of-Thoughts}
\citep{openr1, penedo2025codeforces, lozhkov2025openr1math220k, bercovich2025llamanemotronefficientreasoningmodels}, producing the Qwen3-30B-A3B-ReasoningSFT model. We conduct reinforcement learning experiments on this model.
We compare GSPO and GRPO+R3 with a learning rate of $2\times 10^{-6}$ when $\text{mini\_step}=4$, and compare GRPO and GRPO+R3 with a learning rate of $3\times 10^{-6}$ when $\text{mini\_step}=1$; all other hyperparameters follow the configuration described in Section~\ref{sec:experiments:setting}.

\begin{figure}[h]
\centering
\includegraphics[width=\textwidth]{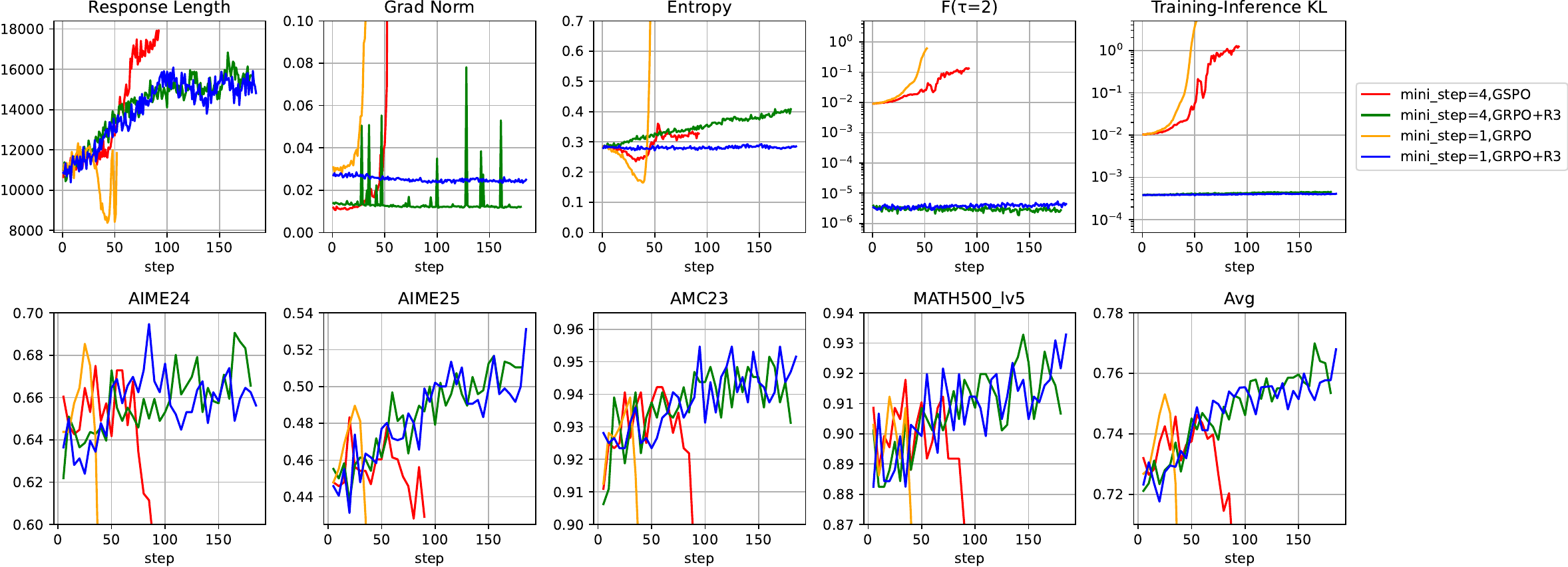}
\caption{The detailed training metrics and evaluation results of the RL training process of Qwen3-30B-A3B-ReasoningSFT}
\label{fig:appendixC}
\end{figure}

The results are shown in Table \ref{table:appendixC}, and training and validation metrics are illustrated in Fig \ref{fig:appendixC}. The experiments show that training processes without R3 collapse (for $\text{mini\_step}=4$, GSPO collapses at training step $90$, and for $\text{mini\_step}=1$, GRPO collapses at training step $40$), while training processes with R3 remain robust. In addition, training with R3 exhibits a more stable pattern of sequence-length growth as well as more stable entropy and gradient norms throughout the training process.
\begin{table}[h]
\centering
\small
\caption{Evaluation results of the RL training process of Qwen3-30B-A3B-ReasoningSFT}
\label{table:appendixC}
\begin{tabular}{lcccccc}
\toprule
& \multicolumn{5}{c}{Best Metric(Best Global Step)} \\
\midrule
Method & AIME24$(\uparrow)$ & AIME25$(\uparrow)$ & AMC23$(\uparrow)$ & MATH500 Lv5$(\uparrow)$ & Avg$(\uparrow)$ & Crash Step \\
\midrule
\rowcolor{gray!30}\multicolumn{7}{c}{\textbf{Qwen3-30B-A3B-ReasoningSFT, mini\_step=4, max\_global\_step=180}} \\
\midrule
GSPO       & 67.50(35) & 48.33(20) & 94.21(60) & 91.79(35) & 74.62(55) & 90 \\
GRPO+R3  & \textbf{69.06}(165) & \textbf{51.66}(155) & \textbf{95.15}(160) & \textbf{93.28}(145) & \textbf{77.00}(165) & - \\
\midrule
\rowcolor{gray!30}\multicolumn{7}{c}{\textbf{Qwen3-30B-A3B-ReasoningSFT, mini\_step=1, max\_global\_step=180}} \\
\midrule
GRPO & 68.54(25) & 48.95(25) & 93.90(30) & 91.23(20) & 75.30(25) & 40 \\
GRPO+R3 & \textbf{69.47}(85) & \textbf{53.12}(180) & \textbf{95.46}(170) & \textbf{93.28}(180) & \textbf{76.79}(180) & - \\
\bottomrule
\end{tabular}
\end{table}

\newpage
\section{Routing-Mask Caching Size Analysis}
\label{sec:appdG}

\subsection{Routing-Mask Cache in Inference Framework Compared with KV Cache}

For the inference framework, the routing-mask cache is stored in CPU memory and is only \(\frac{1}{128}\) the size of the KV cache stored in GPU HBM. Routing masks are maintained alongside KV cache entries, but with a significantly smaller memory footprint due to their lower dimensionality and their storage on CPU memory. The datatype of the KV cache is \texttt{bfloat16} (2 bytes), and the datatype of the routing mask is \texttt{int16} (2 bytes).

Assuming 40 GB of GPU memory allocated for KV cache prefix caching per GPU, only about 2.5 GB of CPU memory per node is required to store routing masks for eight GPUs, which results in negligible deployment overhead. The comparison is summarized in Table~\ref{tab:inference_mask_cache}.

\begin{table}[h]
\centering
\small
\caption{Comparison of KV Cache and Routing Mask Cache in the Inference Framework}
\label{tab:inference_mask_cache}
\resizebox{\textwidth}{!}{%
\begin{tabular}{lcccccc}
\toprule
\multicolumn{1}{l}{} & \multicolumn{1}{c}{KV Cache} & \multicolumn{1}{c}{Routing Mask Cache} \\
\midrule
Each Token & DataType $\times$ \#Layer $\times$ (KDim + VDim) & DataType $\times$ \#Layer $\times$ TopK \\
Each Token in Qwen3-30B-A3B & \(2 \times 48 \times (512 + 512)\) & \(2 \times 48 \times 8\) \\
437K Tokens per GPU & 40 GB on GPU HBM & 0.3 GB on CPU Memory \\
437K $\times$ 8 Tokens per Node (8 GPUs) & \(40\text{ GB} \times 8\) on HBM & 2.5 GB on CPU Memory \\
\bottomrule
\end{tabular}}
\end{table}

\subsection{Routing Mask in Training Framework per Global Training Step}

For the training framework, the maximum memory footprint of routing masks per global step is determined by the batch configuration.
Let the number of prompts be \(B\), the number of groups per prompt be \(G\), the maximum sequence length be \(N\), the number of layers be \(L\), and each layer selects \(K\) experts. The routing mask stores the expert indices of all selected experts using the \texttt{int16} datatype. The maximum routing-mask memory requirement is

\[
\text{Mask Size} = 2 \times B \times G \times N \times L \times K \text{ bytes}.
\]

For the configuration \(B = 256\), \(G = 8\), \(N = 32768\), \(L = 48\), and \(K = 8\), the maximum routing-mask memory footprint is approximately 48 GB.

In practice, these masks are distributed across the training infrastructure rather than being stored on a single machine. They are typically stored in CPU memory and only transferred into GPU memory when required during training.

If the router states are distributed across \(T\) training nodes (e.g., \(T = 32\)), each machine stores only around 1.5 GB of mask data. For modern AI clusters equipped with high-bandwidth interconnects and large memory capacity, both the storage and communication overhead introduced by such routing masks are well within operational limits.

\section{Formal Definition of Router-Discrepancy Metrics}

As analyzed in Section~\ref{sec:3.2}, we have already discussed the qualitative differences in routing behavior.
Below, we formally define the meanings of the x-axis and y-axis in Figure~\ref{fig:321} using mathematical notation.

Let \(D = \{s_1, s_2, \ldots, s_B\}\) denote a batch of sequences.
For each sequence \(s_i\), let its length be \(N_i\), and let the model depth be \(L\).
We define \(d(i, t, l)\) as the difference between the router selections during training and inference at layer \(l\) for token \(s_{i,t}\).

\textbf{Router-Level: } (Figure \ref{fig:321:a})

\[
y = \frac{1}{L \cdot \sum_{i=1}^{B} N_i}
\sum_{i=1}^{B} \sum_{t=1}^{N_i} \sum_{l=1}^{L}
\mathbf{1}[d(i,t,l)=x].
\]

\textbf{Token-Level: } (Figure \ref{fig:321:b})

\[
y = \frac{1}{\sum_{i=1}^{B} N_i}
\sum_{i=1}^{B} \sum_{t=1}^{N_i}
\mathbf{1}\left[\sum_{l=1}^{L} d(i,t,l)=x\right].
\]

\textbf{Sequence-level: } (Figure \ref{fig:321:c})

\[
y = \frac{1}{B}
\sum_{i=1}^{B}
\mathbf{1}\left[
x \le \frac{1}{N_i}
\sum_{t=1}^{N_i}\sum_{l=1}^{L} d(i,t,l) < x+1
\right].
\]

\newpage
\section{Algorithm Details}
\label{appendix:algorithm}

We present the algorithmic pseudocode of R3 for both the multi mini-step setting (Algorithm~\ref{alg:r3_grpo}) and the single mini-step on-policy setting (Algorithm~\ref{alg:r3_grpo_onpolicy}). Note that for clarity, we only highlight the routing replay behavior during the recompute and update stages, omitting many implementation details such as parallel rollout and Dynamic Sampling.

\begin{algorithm}[H]
  \caption{GRPO Training with R3}
  \label{alg:r3_grpo}
  \begin{algorithmic}
    \STATE {\bfseries Input:} Prompt batch $\mathcal{B}$, model parameters $\theta$, number of mini-steps $E$
    \FOR{each global step}
      \STATE \textit{// Stage 1: Inference engine rollout and reward}
      \FOR{each prompt $x \in \mathcal{B}$}
        \STATE Generate response $y \sim \mu_{\theta_{\mathrm{old}}}(\cdot|x)$ via inference engine
        \STATE Cache per-layer routing masks $\{\mathbf{I}_{\mathrm{infer}}^{(l)}\}$ for all tokens
      \ENDFOR
      \STATE Obtain reward $r(x, y)$ for each response via verifier
      \STATE Compute GRPO advantage $\hat{A}$ from group rewards
      \STATE \textit{// Stage 2: Training engine --- recompute old log-probs with R3}
      \STATE Forward pass replaying $\{\mathbf{I}_{\mathrm{infer}}^{(l)}\}$ to obtain $\pi_{\theta_{\mathrm{old}}}(y_t|x, y_{<t})$
      \STATE \textit{// Stage 3: Mini-step policy optimization with R3}
      \FOR{$e = 1$ {\bfseries to} $E$}
        \STATE Sample mini-batch from $\mathcal{B}$
        \STATE Forward pass replaying $\{\mathbf{I}_{\mathrm{infer}}^{(l)}\}$ to obtain $\pi_\theta(y_t|x, y_{<t})$
        \STATE Compute loss $\mathcal{L}$ using $\pi_\theta$, $\pi_{\theta_{\mathrm{old}}}$, $\mu_{\theta_{\mathrm{old}}}$, and $\hat{A}$
        \STATE Backward and update $\theta$
      \ENDFOR
    \ENDFOR
  \end{algorithmic}
\end{algorithm}

\begin{algorithm}[H]
  \caption{On-Policy GRPO Training with R3}
  \label{alg:r3_grpo_onpolicy}
  \begin{algorithmic}
    \STATE {\bfseries Input:} Prompt batch $\mathcal{B}$, model parameters $\theta$
    \FOR{each global step}
      \STATE \textit{// Stage 1: Inference engine rollout and reward}
      \FOR{each prompt $x \in \mathcal{B}$}
        \STATE Generate response $y \sim \mu_{\theta_{\mathrm{old}}}(\cdot|x)$ via inference engine
        \STATE Cache per-layer routing masks $\{\mathbf{I}_{\mathrm{infer}}^{(l)}\}$ for all tokens
      \ENDFOR
      \STATE Obtain reward $r(x, y)$ for each response via verifier
      \STATE Compute GRPO advantage $\hat{A}$ from group rewards
      \STATE \textit{// Stage 2: Training engine --- policy optimization with R3}
      \STATE Forward pass replaying $\{\mathbf{I}_{\mathrm{infer}}^{(l)}\}$ to obtain $\pi_\theta(y_t|x, y_{<t})$
      \STATE Compute loss $\mathcal{L}$ using $\pi_\theta$, $\mu_{\theta_{\mathrm{old}}}$, and $\hat{A}$
      \STATE Backward and update $\theta$
    \ENDFOR
  \end{algorithmic}
\end{algorithm}

\newpage
\section{Dataset Filtering Configuration}

We filtered the math-problem dataset through multiple stages to construct a high-quality subset for programmatic verification and training.

\begin{itemize}
    \item \textbf{Weak-Model Filtering.} We applied \texttt{DeepSeek-R1-Distill-Qwen-7B} \citep{dpskr1} to perform 16 rounds of filtering. Problems with a pass rate greater than 90\% under this weak-model evaluation were removed.
    \item \textbf{Strong-Model Filtering.} We applied the stronger \texttt{DeepSeek-R1-0528} \citep{dpskr1} models for one round of filtering and collected problems that the strong model could not solve.
    \item \textbf{Problem Rewriting.} For problems that failed the strong-model filtering step, we suspect that some failures are due to formatting issues or proof-style requirements. We therefore used several state-of-the-art models (for example, \texttt{DeepSeek-V3})\citep{liu2024deepseek} to attempt automatic rewriting of these problems into forms that can be programmatically verified.
    \item \textbf{Refiltering.} Problems that were rewritten in the previous step were re-evaluated using both the weak-model and strong-model filtering procedures; problems that then passed the filters were removed from the failure set.
\end{itemize}

\section{Problem Verifier Configuration}

The verifier is designed to extract a final parseable mathematical expression from the model output and compare it to the reference solution, supporting both purely numeric answers and symbolic expressions.

\begin{itemize}
    \item \textbf{Output Format.} The model is prompted to perform step-by-step reasoning and to place the final answer inside a boxed expression. The prompt template (presented here as a Python f-string) is:
\begin{verbatim}
f"{problem}\nPlease reason step by step,
and put your final answer within \\boxed{}."
\end{verbatim}
    \item \textbf{Verification Tool.} We use the open-source \texttt{Math-Verify} library\footnote{\url{https://github.com/huggingface/Math-Verify}} to extract the last parseable mathematical expression from the model's answer and compare it against the reference solution. This verifier configuration supports both pure numeric answers and expressions containing variables by comparing semantic mathematical expressions rather than performing raw string matching.
\end{itemize}

\newpage
\section{Token Type Analysis of Routing Discrepancies}
\label{sec:appdL}

We conducted an additional analysis focused on quantifying the routing discrepancy at the token level.

Specifically, we evaluated the Qwen3-30B-A3B model on a set of 2,048 mathematical problems. During inference, we recorded the routing distribution for each generated token. We then fed the same token sequences through the training framework in a forward pass (without R3), again collecting routing distribution for every token.

For each token, we computed its routing discrepancy, defined as the difference in selected experts between the two passes. The discrepancy ranges from 0 to $L \times K$, where $L$ denotes the number of MoE layers and $K$ represents the number of activated experts. Afterward, we filtered out low-frequency tokens from vocabulary (frequency threshold: 0.02\%) and computed the mean discrepancy for the remaining tokens. The top 50 tokens with the highest mean discrepancy are shown in Table \ref{tokentype2}.

\begin{figure}[H]
\centering
\includegraphics[width=0.9\textwidth]{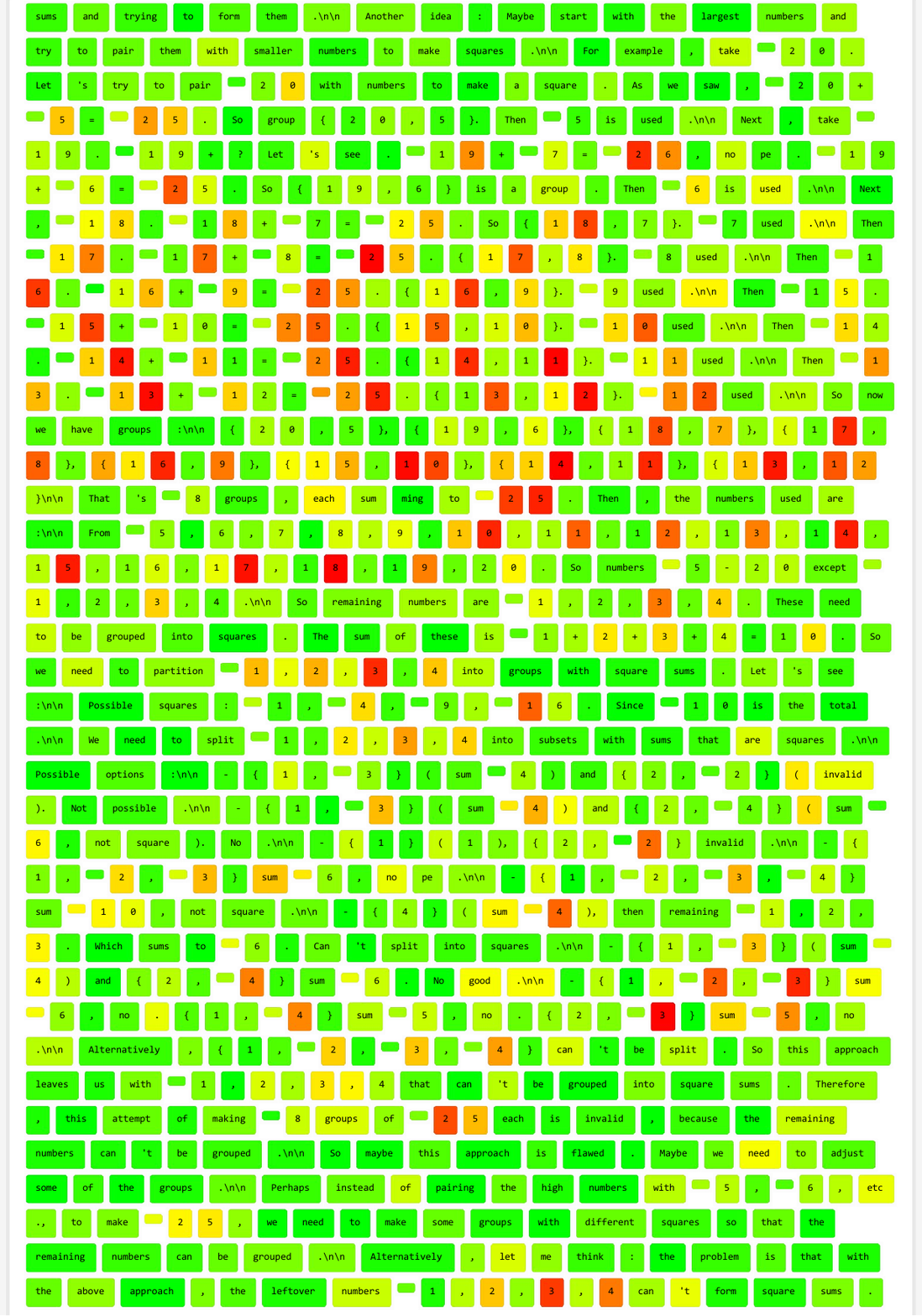}
\label{fig:appendixX}
\caption{Visualization of routing discrepancy for the response segment, where red indicates large routing discrepancy and green indicates small routing discrepancy}
    \label{tokentype1}
\end{figure}

In Figure \ref{tokentype1}, we visualize the token-level routing discrepancy of a response segment, with red indicating greater routing differences.

A clear pattern emerges: tokens corresponding to mathematical computation exhibit substantially higher routing discrepancy compared to others.

One possible explanation is that performing mathematical computation requires the model to simulate complex reasoning steps in the hidden space. During this process, the hidden states change more sharply, which makes the routing decisions less stable and leads to larger differences between training and inference.

In short, the discrepancy does not happen evenly across all tokens. It mainly appears in certain types of tokens, especially those involving mathematical computation, where the difference becomes noticeably larger.

\begin{table}[H]
    \centering
    \small
    \caption{The top 50 tokens ranked by routing discrepancy}
    \label{tokentype2}
    \begin{tabular}{|r|r|l|r|}
    \hline
    \multicolumn{1}{|c|}{Rank} & \multicolumn{1}{c|}{Token ID} & \multicolumn{1}{c|}{Token String} & \multicolumn{1}{c|}{Mean Token-Level Diff Router} \\
    \hline
    1 & 15 & \texttt{"0"} &                                       16.9834                                       \\ \hline
    2 & 37018 & \texttt{"frac"} &                                 16.6235                                 \\ \hline
    3 & 248 & \texttt{(unvisualizable)} &                   16.5253                \\ \hline
    4 & 50853 & \texttt{"cdot"} &                                 15.9350                                 \\ \hline
    5 & 230 & \texttt{(unvisualizable)} &                   15.8782                \\ \hline
    6 & 20 & \texttt{"5"} &                                       15.7010                                       \\ \hline
    7 & 116 & \texttt{(unvisualizable)} &                   15.6420                \\ \hline
    8 & 18 & \texttt{"3"} &                                       15.6307                                       \\ \hline
    9 & 23 & \texttt{"8"} &                                       15.5839                                       \\ \hline
    10 & 19 & \texttt{"4"} &                                      15.3182                                      \\ \hline
    11 & 21 & \texttt{"6"} &                                      15.2691                                      \\ \hline
    12 & 17 & \texttt{"2"} &                                      15.2680                                      \\ \hline
    13 & 22 & \texttt{"7"} &                                      15.0927                                      \\ \hline
    14 & 3245 & \texttt{" least"} &                               15.0805                               \\ \hline
    15 & 16 & \texttt{"1"} &                                      14.8400                                      \\ \hline
    16 & 94 & \texttt{(unvisualizable)} &                   14.6765                \\ \hline
    17 & 258 & \texttt{"in"} &                                    14.4592                                    \\ \hline
    18 & 15896 & \texttt{(unvisualizable)} &                14.3470             \\ \hline
    19 & 24 & \texttt{"9"} &                                      14.3465                                      \\ \hline
    20 & 104 & \texttt{(unvisualizable)} &                  13.2339               \\ \hline
    21 & 79075 & \texttt{"boxed"} &                               12.9715                               \\ \hline
    22 & 944 & \texttt{"'t"} &                                    12.6392                                    \\ \hline
    23 & 31807 & \texttt{"$_{2}$"} &                              12.6165                                   \\ \hline
    24 & 26888 & \texttt{"sqrt"} &                                11.9509                                \\ \hline
    25 & 15940 & \texttt{"sin"} &                                 11.5561                                 \\ \hline
    26 & 15170 & \texttt{"\}\{"} &                                11.4702             \\ \hline
    27 & 35702 & \texttt{"\{\textbackslash\textbackslash"} &       9.8007             \\ \hline
    28 & 9407 & \texttt{"cos"} &                                   9.6640             \\ \hline
    29 & 14 & \texttt{"/"} &                                       8.7482             \\ \hline
    30 & 72 & \texttt{"i"} &                                       8.7010            \\ \hline
    31 & 62 & \texttt{"\_"} &                                      8.5313             \\ \hline
    32 & 272 & \texttt{" c"} &                                     8.4916             \\ \hline
    33 & 90 & \texttt{"\{"} &                                      8.4155             \\ \hline
    34 & 89 & \texttt{"z"} &                                       8.1327             \\ \hline
    35 & 33 & \texttt{"B"} &                                       8.0483             \\ \hline
    36 & 2086 & \texttt{" second"} &                               7.9709              \\ \hline
    37 & 11067 & \texttt{" sides"} &                               7.8640             \\ \hline
    38 & 77 & \texttt{"n"} &                                       7.8423            \\ \hline
    39 & 34 & \texttt{"C"} &                                       7.7431             \\ \hline
    40 & 88 & \texttt{"y"} &                                       7.6464             \\ \hline
    41 & 89669 & \texttt{"$_{0}$"} &                               7.5362             \\ \hline
    42 & 48245 & \texttt{"$\pi$"} &                                7.5002             \\ \hline
    43 & 12 & \texttt{"-"} &                                       7.4991             \\ \hline
    44 & 65 & \texttt{"b"} &                                       7.4563             \\ \hline
    45 & 10 & \texttt{"+"} &                                       7.4212             \\ \hline
    46 & 1242 & \texttt{"sum"} &                                   7.4100              \\ \hline
    47 & 425 & \texttt{" B"} &                                     7.3921              \\ \hline
    48 & 74 & \texttt{"k"} &                                       7.3901             \\ \hline
    49 & 30986 & \texttt{"$_{1}$"} &                               7.3890             \\ \hline
    50 & 198 & \texttt{"\textbackslash n"} &                       7.3686             \\ \hline
    \end{tabular}
\end{table}

\newpage
\section{Per-Layer Routing Divergence between Training and Inference}
\label{sec:layerwise_routing}

\begin{figure}[h]
\centering
\includegraphics[width=0.8\linewidth]{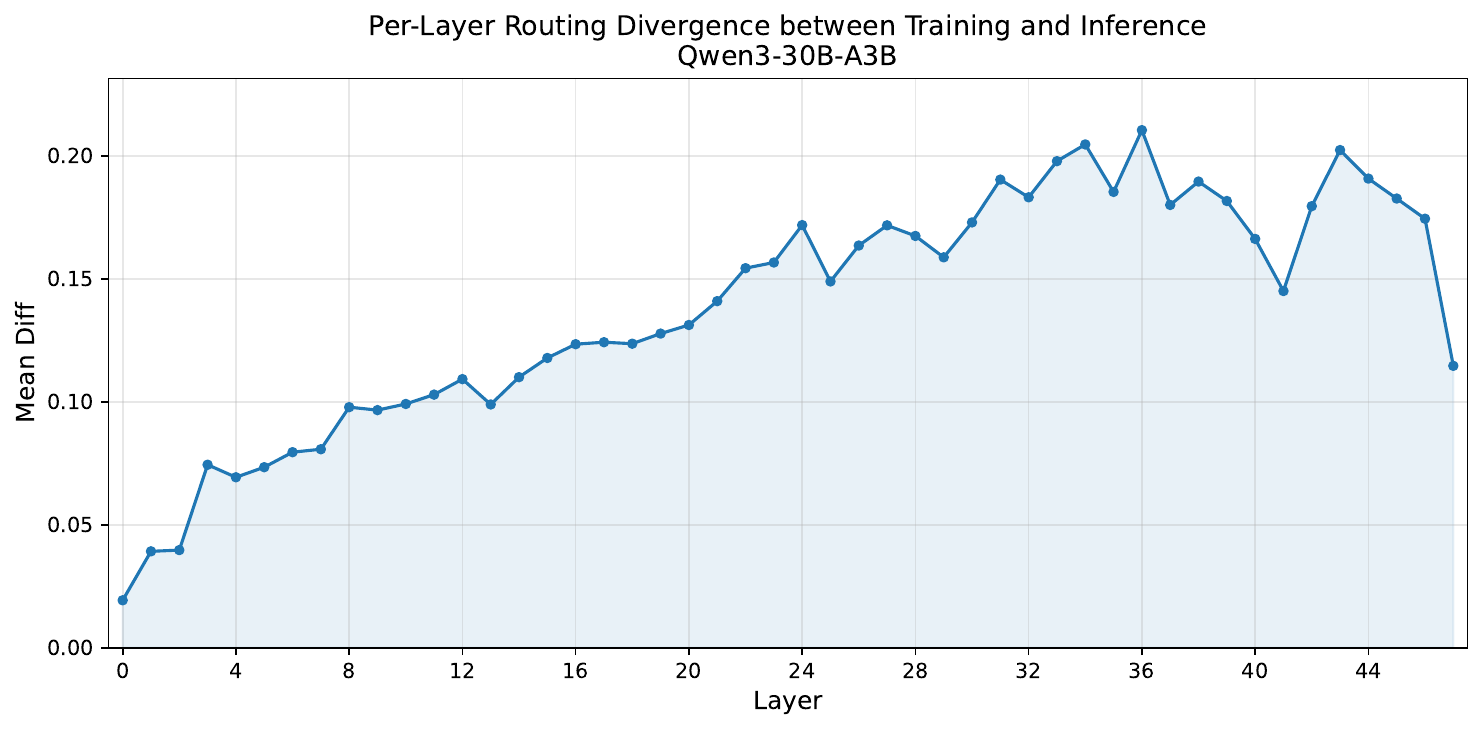}
\caption{Per-layer routing divergence (Mean Diff) between training and
inference for Qwen3-30B-A3B across all 48 layers.}
\label{fig:layerwise_routing}
\end{figure}

To better understand how the routing inconsistency flows across layers,
we measure the routing difference at every layer of Qwen3-30B-A3B.
For each layer, we compute the mean difference (\emph{Mean Diff}) between the
routing distributions during training and inference.
The results are shown in Figure~\ref{fig:layerwise_routing}.

As shown in Figure~\ref{fig:layerwise_routing}, the divergence is small in the
shallow layers ($0.0193$ at layer~0), grows with depth, reaches its peak at
layer~36 ($0.2104$), and then drops in the last few layers.

\section{Influence of Model Size on Training-Inference Discrepancy}
\label{sec:model_size}

We examine whether model size affects the training-inference discrepancy.
We evaluate Qwen3 models of sizes 0.6B, 1.7B, 4B, 8B, 32B, 30B-A3B, and
30B-A3B + R3, including both Thinking and Base variants.
The results are shown in Table~\ref{tab:model_size}.

The results show that: (1) no clear pattern emerges between model size and the
discrepancy among dense models; (2) the MoE model (30B-A3B) shows larger
discrepancy than all dense models; (3) applying R3 reduces the MoE discrepancy
to levels comparable with dense models.

\begin{table}[h]
\centering
\caption{Training-inference discrepancy across different model sizes.
KL denotes KL divergence and $F(\tau{=}2)$ denotes the F metric with
threshold $\tau=2$.}
\label{tab:model_size}
\small
\setlength{\tabcolsep}{5pt}
\begin{tabular}{lcccc}
\toprule
Model & KL (Base) & KL (Thinking) & $F(\tau{=}2)$ (Base) & $F(\tau{=}2)$ (Thinking) \\
\midrule
0.6B        & 6.631e-04 & 7.549e-04 & 4.294e-06 & 1.767e-05 \\
1.7B        & 1.028e-03 & 9.389e-04 & 6.371e-06 & 6.185e-05 \\
4B          & 6.037e-04 & 6.446e-04 & 1.868e-05 & 1.066e-05 \\
8B          & 9.834e-04 & 6.471e-04 & 1.575e-05 & 8.651e-06 \\
32B         & --        & 8.709e-04 & --        & 1.428e-05 \\
\midrule
\textbf{30B-A3B}     & \textbf{1.339e-03} & \textbf{1.548e-03} & \textbf{2.992e-04} & \textbf{3.377e-04} \\
30B-A3B + R3 & 5.303e-04 & 7.347e-04 & 8.860e-06 & 1.270e-05 \\
\bottomrule
\end{tabular}
\end{table}

\section{Training Hyperparameters}
\begin{table}[h]
\caption{Default Training hyperparameters of Reinforcement Learning.}
\centering
\small
\begin{tabular}{lc}
\toprule
\textbf{Hyperparameter} & \textbf{Value} \\
\midrule
Batch Size & 256 \\
Group Size & 8 \\
Mini Step & 8 or 1 \\
Max Prompt Length & 2048 \\
Max Response Length & 30720 \\
\midrule
Optimizer & AdamW \\
Weight Decay & 0.1 \\
Learning Rate & $3 \times 10^{-6}$ or $1 \times 10^{-6}$ \\
Learning Rate Scheduler & Constant \\
Beta1 & 0.9 \\
Beta2 & 0.999 \\
Epsilon & $1 \times 10^{-8}$ \\
Max Gradient Norm & 1.0 \\
\midrule
DAPO Clip-low & 0.20 \\
DAPO Clip-high & 0.27 \\
GSPO Clip-low & 3e-4 \\
GSPO Clip-high & 4e-4 \\
TIS Clip & 2.0 \\
\bottomrule
\end{tabular}
\end{table}

\end{document}